\documentclass[final]{siamltex}
\usepackage{verbatim}
\pdfoutput=1
\usepackage{graphicx, epsfig}
\usepackage{float}
\usepackage{amsmath,amssymb}
\usepackage{hyperref}
\usepackage{fancyref}
\usepackage{xcolor}
\usepackage{subfig}
\usepackage{multirow}
\usepackage{bbm}
\usepackage{float}
\usepackage{soul}
\usepackage[makeroom]{cancel}
\usepackage[normalem]{ulem}
\usepackage{bm}
\usepackage{tikz}
\usepackage{algorithm}
\usepackage{algpseudocode}
\usepackage{comment}

\title{uncertainty quantification for DeepONets with Ensemble Kalman Inversion}
\author{Andrew Pensoneault\thanks{Department of Mathematics, University of Iowa, Iowa City, IA 52246, USA. Email: 
andrew-pensoneault@uiowa.edu.}
\and  Xueyu Zhu\thanks{Department of Mathematics, University of Iowa, Iowa City, IA 52246. USA. Email: xueyu-zhu@uiowa.edu.}
}

\begin{document}
\maketitle

\begin{abstract}
In recent years, operator learning, particularly the DeepONet, has received much attention for efficiently learning complex mappings between input and output functions across diverse fields. However, in practical scenarios with limited and noisy data, accessing the uncertainty in DeepONet predictions becomes essential, especially in mission-critical or safety-critical applications. Existing methods, either computationally intensive or yielding unsatisfactory uncertainty quantification, leave room for developing efficient and informative uncertainty quantification (UQ) techniques tailored for DeepONets. In this work, we proposed a novel inference approach for efficient UQ for operator learning by harnessing the power of the Ensemble Kalman Inversion (EKI) approach. EKI, known for its derivative-free, noise-robust, and highly parallelizable feature, has demonstrated its advantages for UQ for physics-informed neural networks \cite{Pensoneault2023}.
Our innovative application of EKI enables us to efficiently train ensembles of DeepONets while obtaining informative uncertainty estimates for the output of interest. We deploy a mini-batch variant of EKI to accommodate larger datasets, mitigating the computational demand due to large datasets during the training stage. Furthermore, we introduce a heuristic method to estimate the artificial dynamics covariance, thereby improving our uncertainty estimates.
Finally, we demonstrate the effectiveness and versatility of our proposed methodology across various benchmark problems, showcasing its potential to address the pressing challenges of uncertainty quantification in DeepONets, especially for practical applications with limited and noisy data.
\end{abstract}


\pagestyle{myheadings}
\thispagestyle{plain}

\section{Introduction}
\label{sec:Intro}

In recent years, deep learning has played a pivotal role in the development of data-driven solutions for a wide class
of scientific and engineering challenges \cite{Lu_2021, RAISSI2019686, kovachki2022neural, Kobyzev_2021, chen2019neural}. One area of research that has captured significant interest is  {\it operator learning}, involving learning the unknown mapping between input and output functions. This problem is ubiquitous across various disciplines \cite{kovachki2022neural, Lu_2021}. Several approaches have been developed along this line, including Fourier Neural Operators  DeepONets \cite{Lu_2021}, \cite{li2021fourier}, Laplace Neural Operators \cite{cao2023lno}, and Wavelet Neural Operators \cite{TRIPURA2023115783}.

One particularly popular approach in operator learning is DeepONets  \cite{Lu_2021}. DeepONets are built upon the foundation of the universal approximation theorem for nonlinear operators in neural networks \cite{chen1995universal}. The architecture consists of two networks: the ``branch" network encodes input function information, while the ``trunk" network encodes the locations where
one evaluates the output function. This architecture enables DeepONets to efficiently learn nonlinear
operators given available data. The potential of DeepONets has been demonstrated in a variety of applications, such as estimating boundary layers \cite{Leoni_2021}, predicting multiscale bubble growth dynamics \cite{Lin_2021}, and simulating high-velocity flow in chemically-reactive fluids \cite{MAO2021110698}. 
 
In scenarios with noise and limited data, the assessment of uncertainty's impact on DeepONet predictions becomes crucial. This is especially true for real-world applications where reliability is important. In the context of Bayesian methods, various approaches to UQ have been developed for DeepONets. Notable among these are the HMC \cite{zou2023uncertainty} and its variants\cite{lin2021accelerated,moya2022deeponet}. While Hamiltonian Monte Carlo (HMC) methods are popular for Bayesian inference, they are computationally expensive and may not scale well for large architectures and with massive datasets, such as is the case with DeepONets \cite{yang2022scalable}.  Variational Inference (VI) serves as an alternative approach within DeepONets. VI addresses a deterministic optimization problem over a family of parameterized densities to approximate the posterior distribution of the subject quantity \cite{Blei_2017}. 
Existing work includes VB-DeepONet \cite{Shailesh_2023} and Prob-DeepONet \cite{moya2022deeponet}.  However, the quality of VI estimates depends on the choice of parameterized distribution space  \cite{yang2022scalable}.  
In the context of non-Bayesian methods, UQDeepONet has recently been proposed  \cite{yang2022scalable}. This approach employs randomized prior networks within a deep ensemble framework. These ensembles are efficient but require careful parameter tuning, particularly concerning the weight given to a fixed randomly initialized random prior network in each ensemble member.  Other related UQ research for DeepOnets can be found in \cite{guo2023ib, zou2023uncertainty}. In light of these challenges, it is highly desirable to devise innovative methods that effectively address the tradeoffs between efficiency and quality in UQ  for DeepONets, especially in the presence of noisy and limited data. 

To tackle these challenges, we adopt the Ensemble Kalman Inversion (EKI) approach \cite{Kovachki_2019, Iglesias_2013}, an ensemble-based method to solve the inverse
problems, to effectively quantify the uncertainties of DeepONets, given the noises in the output measurements. EKI offers numerous advantages such as being derivative-free, noise-robust, highly parallelizable, and well-suited for high-dimensional parameter inference \cite{LopezGomez_2022}. Originally devised for solving inverse problems in the context of optimization\cite{Kovachki_2019}, EKI was later extended as a derivative-free optimization technique for neural network training. However, its potential for uncertainty quantification has not been thoroughly explored. Recent successful applications of EKI in training Physics Informed Neural Networks (PINNs) \cite{Pensoneault2023} have demonstrated its efficacy in enabling efficient high-dimensional UQ. Encouraged by these advancements, we propose a novel approach for UQ in DeepONets, capitalizing on EKI's advantageous properties. It's crucial to recognize that DeepONets typically involve larger network architectures and substantially larger datasets compared to PINNs. To accommodate larger datasets, we utilize a minibatch version of EKI \cite{Kovachki_2019} to mitigate the computational demand during the training phase. Additionally, we introduce a heuristic method to estimate the covariance matrix of artificial dynamics based on available data. This technique aims to prevent ensemble collapse while also minimizing excessive parameter perturbation during the model's training. To the best of our knowledge, this work represents the {\it first attempt} to apply EKI for informative uncertainty quantification for DeepONets.

The rest of the paper is organized as follows:  in Section \ref{sec:bg}, we first briefly introduce operator learning and the DeepONet. We then introduce the framework of the Bayesian DeepONet. Next, we introduce the Ensemble Kalman Inversion as a method for uncertainty quantification. We then introduce the EKI-based Bayesian DeepONet. We show the effectiveness of the EKI for several numerical examples in Section \ref{sec:NumExample}. Finally, we conclude in section \ref{sec:summary}.
\section{Problem Setup and Background}
\label{sec:bg}

Operator learning is a fundamental problem arising in many different fields involving learning an unknown operator that maps input functions to output functions. Many efforts have been developed for efficient  operator learning, such as DeepONet \cite{Lu_2021} and Fourier Neural Operator \cite{li2021fourier}. In this paper, we will focus on the setting of  DeepONet \cite{Lu_2021}. In this section, we introduce the problem setup for operator learning and briefly review DeepONets and Bayesian DeepONets (B-DeepONets) \cite{Lu_2021,lin2021accelerated}.

\subsection{DeepONets}
\label{subsec:problem}
We denote $x\in\mathcal{X}$ ``sensors locations" and $y\in\mathcal{Y}$ ``query locations", where  $\mathcal{X}\subset\mathbb{R}^{d_x}$ and $\mathcal{Y}\subset\mathbb{R}^{d_y}$. Additionally, we use $C(\mathcal{X};\mathbb{R}^{d_u})$ and $C(\mathcal{Y};\mathbb{R}^{d_s})$ to denote the spaces of the continuous input  function $u:\mathcal{X}\to\mathbb{R}^{d_u}$ and the output function $s$: $\mathcal{Y}\to\mathbb{R}^{d_s}$, respectively. 

The goal of operator learning is to learn an operator $\mathcal{G}_\theta:C(\mathcal{X};\mathbb{R}^{d_u})\to C(\mathcal{Y};\mathbb{R}^{d_s})$ from a input/output function pair $\{u^l, s^l\}_{l=1}^{N}$ generated from a unknown ground truth operator $\mathcal{G}:C(\mathcal{X};\mathbb{R}^{d_u})\to C(\mathcal{Y};\mathbb{R}^{d_s})$, where $u^l\in C(\mathcal{X};\mathbb{R}^{d_u})$, $s^l\in C(\mathcal{Y};\mathbb{R}^{d_s})$, and $l=1,...,N$, such that:
\begin{align}
    s^l = \mathcal{G}(u^l)\approx\mathcal{G}_\theta(u^l).
\end{align}

The foundation of the DeepONet can be traced back to the universal approximation theorem, as initially formulated in \cite{chen1995universal}, further generalized in \cite{Lu_2021}. The DeepONet constructs a parameterized operator $\mathcal{G}_\theta(u)(y)$ using two distinct neural networks: a branch network $b:\mathbb{R}^{m\times d_u}\to\mathbb{R}^{n\times d_s}$ that operates on input functions $u$, measured at $m$ fixed sensor locations and outputs a $d_s$-dimensional feature vector; and a trunk network $t:\mathbb{R}^{1\times d_u}\to\mathbb{R}^{n\times d_s}$ that operates on a query location $y$ and outputs a $d_s$-dimensional feature vector.  Given an input function $u$ and query location $y$,   DeepONets predict the value of the corresponding output function $s$ by merging the two $d_s$-dimension feature together:
$$G(u)(y)\approx\mathcal{G}_\theta(u)(y)=\sum_{i=1}^n b_i(u)\cdot t_i(y).$$

Given a data set consisting of $N$ input/output function pairs $\{u^l,s^l\}_{l=1}^N$, and a corresponding set of $P$ query locations $\{y_j^l\}_{j=1}^P$ for output function $s^l$, DeepONets  are traditionally trained via minimization of the empirical risk minimization loss \cite{yang2022scalable}:
\begin{align*}
\mathcal{L}(\theta) &= \frac{1}{NP}\sum_{l=1}^N\sum_{j=1}^P|\mathcal{G}_\theta(u^l)(y_j^l)-s^l(y_j^l)|^2.
\end{align*}

Once trained, DeepONets provide a deterministic prediction of a target output function $s^*$ at a query location $y^*$. In many applications, particularly where the reliability of the prediction is needed, understanding the uncertainty in the system is paramount. Several approaches to provide uncertainty quantification (UQ) in the setting of DeepONets have been explored \cite{yang2022scalable,lin2021accelerated,yang2022scalable,guo2023ib}. In the next section, we shall review a Bayesian version of the DeepONet (B-DeepONet), and based on this to frame a novel Ensemble-based UQ approach to DeepONets.

\begin{figure}
\centering
\includegraphics[scale=0.5]{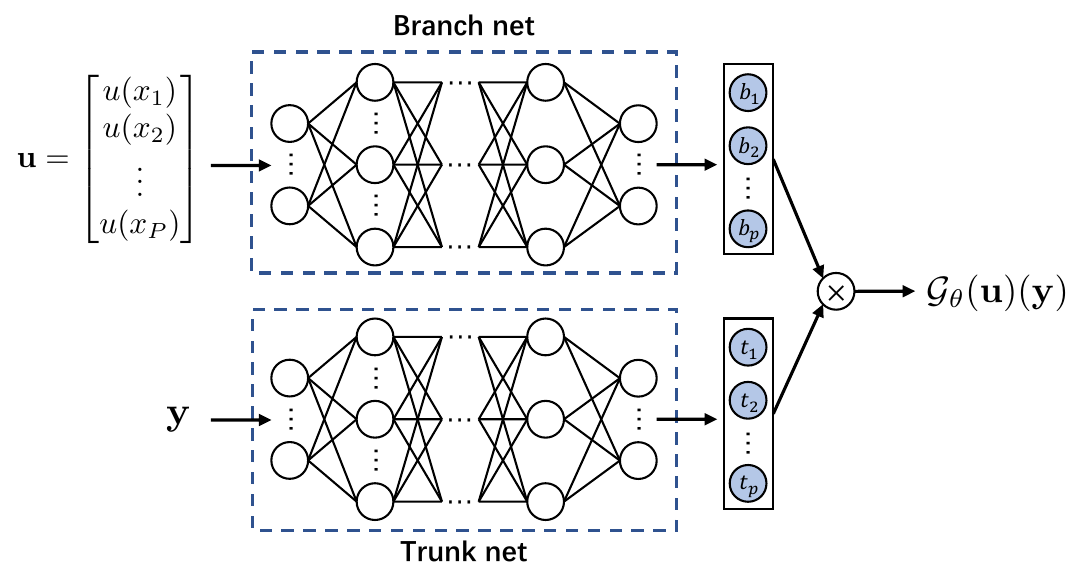}
\caption{Schematic of the DeepONet architecture. The "branch" network, denoted as \(b(u)\), encodes input functions evaluated at predetermined "sensor" locations. The "trunk" network, \(t(y)\), encodes a "query" location where the output function is to be evaluated. DeepONet fuses the encoded information from both networks via a dot product to produce an approximate function output at the ``query" location.}
\label{fig:nn}
\end{figure}

\subsection{Bayesian DeepONet (B-DeepONet)} 
\label{subsec:bdeeponet}
In this work, we only consider the scenarios where the output measurements are noisy and limited, i.e., assuming the training output samples, $G_k^l$, obtained from the ground-truth operator $\mathcal{G}(u^l)(y_k^l)$, are corrupted with independent zero-mean Gaussian noise, i.e.,
\begin{align}
G_j^l = \mathcal{G}(u^l)(y^l_j) + \epsilon_{k}^l,\quad \epsilon_{k}^l\sim\mathcal{N}(0,\sigma_l).
\end{align}
Define the noisy output dataset corresponding to the $l$-th input function $u^l$,  the corresponding  output measurements $\mathcal{D}^l=\{G_k^l\}_{k=1}^P$ at $P$ query locations, and the total noisy dataset $\mathcal{D} = \{\mathcal{D}^l\}_{l=1}^N$. We assume independent and identically distributed (i.i.d.) distribution for data and Gaussian likelihood:
\begin{align}
    p(\mathcal{D}|\theta) &=\prod_{k=1}^{P}\prod_{l=1}^{N}\frac{1}{\sqrt{2\pi\sigma_l^2}}\exp\left(-\frac{\left(G_j^l-\mathcal{G}_\theta(u^l)(y^l_k)\right)^2}{2\sigma_l^2}\right) \label{al:likelihood2}.
\end{align}
In this setting, we assume $\sigma_l>0$ is the standard deviation of the observation error for each input function $u^l$, which is known a priori. Given a prior $p(\theta)$, Bayes' theorem states the posterior pdf $p(\theta|\mathcal{D})$ takes the form 
\begin{align}
    p(\theta|\mathcal{D})\propto p(\theta)p(\mathcal{D}|\theta).
\end{align}

For most choices of priors on $\theta$, computing the posterior distribution is intractable; thus, approximate inference methods are often employed. Given the high dimensionality of neural networks due to a large number of parameters, traditional sampling-based Bayesian inference methods become impractical for these applications \cite{yang2022scalable} due to the sampling efficiency and the cost of gradient-based optimization.  

Alternatively, the EKI is typically employed as an ensemble-based zeroth order optimizer method typically utilized to solve a variety of inverse problems \cite{Kovachki_2019,Guthweissman_2020, liu2023dropout}. EKI offers several advantages: it is derivative-free, noise-robust, highly parallelizable, and requires a smaller number of ensemble members relative to the dimensionality of the parameter space \cite{LopezGomez_2022}. In addition to being utilized as a derivative-free optimizer, the EKI has more recently been used to obtain informative uncertainty estimates in the framework of Bayesian Physics-Informed Neural Networks (B-PINNs) \cite{Pensoneault2023}. Motivated by the success of EKI in the B-PINN setting, we aim to leverage this methodology to quantify uncertainty in DeepONets efficiently. Next, we shall first introduce EKI in the broader context of Bayesian inverse problems to set the stage for our contributions.

\subsection{Ensemble Kalman Inversion (EKI)}
\label{subsec:EKI}

Ensemble Kalman Inversion (EKI) \cite{Iglesias_2013, iglesias_2015, Iglesias_2016, Kovachki_2019} is a class of methods that utilize the Ensemble Kalman Filter (EnKF) \cite{evensen2003ensemble} to solve traditional inverse problems. 
For this paper, we choose to adapt one variant of  EKI methods described in \cite{Pensoneault2023}. 

We assume parameters $\theta\in\mathbb{R}^{N_\theta}$ have a prior distribution $p(\theta)$ and the observations $y\in\mathbb{R}^{N_y}$ are related to the parameters through the observation operator $\mathcal{H}$:
\begin{align}
y = \mathcal{H}(\theta) + \eta\label{al:bip_1},
\end{align}
where $\eta\in\mathbb{R}^{N_y}$ is the measurement error or noise term, following a zero-mean Gaussian distribution with covariance matrix $R\in\mathbb{R}^{N_y\times N_y}$, i.e., ${\eta}\sim\mathcal{N}(0,R)$. From Bayes' theorem, this leads to the following posterior distribution: 
\begin{align}
p({\theta}|{y}) &\propto p({\theta})\exp\left(-\frac{\left\|R^{-1/2}({y}-\mathcal{H}({\theta}))\right\|_2^2}{2}\right)\label{al:bip}.
\end{align}
By assuming a prior of the form $p({\theta})\sim\mathcal{N}({\theta}_0,C_0)$, the posterior becomes
\begin{align}
p({\theta}|{y}) \propto \exp\left(-\frac{\left\|C_0^{-1/2}({\theta}_0-{\theta})\right\|_2^2 + \Big\|R^{-1/2}({y}-\mathcal{H}(\theta))\Big\|_2^2}{2}\right)
\label{al:bip_2}.
\end{align}

EKI approaches based on artificial observations can be framed as state-space models, as in the traditional EnKF setting. The artificial dynamics state-space model formulation corresponding to the original Bayesian inverse problem takes the form \eqref{al:bip_1}: 
\begin{align}
\theta_i&=\theta_{i-1} + {\epsilon}_i,\quad {\epsilon}_i\sim \mathcal{N}(0,Q),\label{al:ad1}\\
{y}_i&=\mathcal{H}(\theta_{i}) + {\eta}_i,\quad {\eta}_i\sim \mathcal{N}(0,R),\label{al:ad2}
\end{align}
where ${\epsilon}_i$ is an artificial noise term with the  covariance $Q\in\mathbb{R}^{N_\theta\times N_\theta}$ and ${\eta}_i$ represents the observation error/noise with the  covariance $R\in\mathbb{R}^{N_y\times N_y}$.  $i$ represents the time index for the  $i$-th EKI iteration. Specifically, given this formulation of the problem, one can apply the EnKF update equations as in \cite{houtekamer2001sequential}. Given an initial ensemble of $J$ ensemble members $\{{\theta}_0^{(j)}\}_{j=1}^J$, the iterative EKI methods iteratively correct the ensemble $\{{\theta}_i^{(j)}\}_{j=1}^J$:
\begin{align}
\hat{\theta}_i^{(j)} &= {\theta}_{i-1}^{(j)} + {\epsilon}_i^{(j)}, \quad {\epsilon}_i^{(j)}\sim \mathcal{N}(0,Q),\label{al:EKI-2-1}\\
\hat{y}^{(j)}_i&=\mathcal{H}(\hat{\theta}_i^{(j)}),\label{al:EKI-2-2}\\
{\theta}_i^{(j)} &= \hat{\theta}_i^{(j)} + {C}_{i}^{\hat{\theta} y}({C}^{yy}_i + R)^{-1}(y-y^{(j)}_{i}+{\eta}_i^{(j)}), \quad {\eta}_i^{(j)}\sim \mathcal{N}(0,R),\label{al:EKI-2-3}
\end{align}
where ${C}^{yy}_i$ and ${C}^{\hat{\theta}y}_i$ are the sample covariance matrices defined as follows: 
\begin{align}
{C}^{yy}_i &= Y_iY_i^T,\quad
{C}^{\hat{\theta}y}_i = \Theta_i Y_i^T,\label{al:EKI-2-5}\\
Y_i &= \frac{1}{\sqrt{J-1}} [\hat{y}_{i}^{(1)}-\bar{y}_i,\hdots,\hat{y}_{i}^{(J)}-\bar{y}_i],\label{al:EKI-2-6}\\
\Theta_i &= \frac{1}{\sqrt{J-1}} [\hat{\theta}_{i}^{(1)}-\bar{\theta}_i,\hdots,\hat{\theta}_{i}^{(J)}-\bar{\theta}_i].\label{al:EKI-2-7}
\end{align} 
Here, $\bar{\theta}_i$ and $\bar{{y}}_i$ are the corresponding sample average of prior ensembles $\{\hat{\theta}^{(j)}_i\}$ and $\{\hat{{{y}}}^{(j)}_i\}$ at $j$-th iteration.

\begin{algorithm}
\caption{Ensemble Kalman Inversion (EKI)}\label{alg:eki}
\begin{algorithmic}[1]
\State Input: $y$ (observations), $Q$ (artificial dynamics covariance), $R$ (observation covariance)
\State Initialize prior samples for $j=1,\hdots,J$:
\begin{align*}
\theta_0^{(j)}&\sim P(\theta_0). 
\end{align*}
\For{$i=1,\hdots,I$}
\State Obtain $i$-th prior parameter and measurement ensembles for $j=1,\hdots,J$:
\begin{align*}
{\epsilon}_i^{(j)}&\sim \mathcal{N}(0,Q), \quad
{\eta}_i^{(j)}\sim \mathcal{N}(0,R).\\
\hat{\theta}_i^{(j)}&={\theta}_i^{(j)}+{\epsilon}_i^{(j)}.\\
\hat{y}_i^{(j)}&=\mathcal{H}(\hat{\theta}_i^{(j)}).
\end{align*}
\State Compute the sample mean and covariance:
\begin{align*}
\bar{\theta}_i &= \frac{1}{J}\sum_{j=1}^J \hat{\theta}_{i}^{(j)}, \quad
\bar{y}_i = \frac{1}{J}\sum_{j=1}^J \hat{y}_{i}^{(j)}.\\
{C}^{\hat{\theta}y}_i &= \frac{1}{J-1}\sum_{j=1}^J (\hat{\theta}^{(j)}_{i}-\bar{\theta}_i)(\hat{y}^{(j)}_{i}-\bar{y}_i)^T.\\
{C}^{yy}_i &= \frac{1}{J-1}\sum_{j=1}^J (\hat{y}^{(j)}_{i}-\bar{y}_i)(\hat{y}^{(j)}_{i}-\bar{y}_i)^T.
\end{align*}
\State Update the posterior ensemble for $j=1,\hdots,J$: 
\begin{align*}
{\theta}_i^{(j)} &= \hat{\theta}_i^{(j)} + C^{\hat{\theta} y}({C}^{yy}_i + R)^{-1}(y-\hat{y}_i^{(j)} + {\eta}_i^{(j)}).
\end{align*}
\EndFor
\State Return: ${\theta}_I^{(1)},\hdots,{\theta}_I^{(J)}$
\end{algorithmic}
\end{algorithm}

\subsection{EKI-Bayesian DeepOnet}
\label{subsec:EKI-B-DO} 
Motivated by the success of the EKI  for efficient inference in the Bayesian physics-informed neural networks (B-PINNs) context presented in \cite{Pensoneault2023}, we propose a novel approach called EKI B-DeepONet that utilizes EKI for efficient inference, benefiting from EKI's gradient-free nature, embarrassingly parallelism, and scalability, particularly advantageous in high-dimensional parameter spaces as encountered in the case of B-DeepONet.
 
To apply EKI, we recall the notation of B-DeepOnet discussed in \ref{subsec:bdeeponet} under the lens of EKI setup.  The corresponding observations $\mathbf{y}$, observation operator $\mathcal{H}(\theta)$, and observation covariance $R$ is defined as follows:
\begin{align}
    \mathbf{y} &=
        \mathcal{D} =\{\mathcal{D}^l\}_{l=1}^N, \quad \mathcal{D}^l=\{G_k^l\}_{k=1}^P, \\
    \mathcal{H}_l(\theta)&=
       [\mathcal{G}_\theta(u^l)(y_1^l), \mathcal{G}_\theta(u^l)(y_2^l), \hdots, \mathcal{G}_\theta(u^l)(y_P^l)],  \\
       \mathcal{H}(\theta) &= [\mathcal{H}_1(\theta), \mathcal{H}_2(\theta), \hdots, \mathcal{H}_N(\theta)] \label{al:H_op},\\
       R_l &= \sigma_l^2 I_P,\\
       R &= diag(R_1, R_2,\hdots, R_N) \label{al:R_def},
\end{align}

where $I_P$ is a $P\times P$ dimensional identity matrix. Note that as $\mathcal{G}_\theta$ is a neural network, the forward evaluations are inexpensive to evaluate, and thus the observation operator $\mathcal{H}$ is efficient to evaluate even for large ensemble sizes. This is not the case in traditional data assimilation settings, where $\mathcal{H}$ is usually a composition of the forward solver of the underlying PDE and observation operator. 

In this particular scenario, it is worth noting that a direct application of the EKI over the full dataset will result in a cubic growth of computational complexity with respect to the total data dimension, $NP$. To mitigate this challenge, we adopt a mini-batch strategy for EKI, as previously suggested in \cite{Kovachki_2019}. This strategy involves dividing the dataset into batches of size $m_t$, effectively managing the complexity relative to the batch size $m_t$ during each iteration. At each EKI iteration, samples are drawn without replacement from the entire set of observations $\mathbf{y}$, and corresponding samples from the operator $\mathcal{H}(\theta)$ and covariance matrix $R$ are generated accordingly.

After inference of the parameters of the neural network, given a new input function $u^*$ and a query location $y^*$, the ensemble mean and standard deviation for the B-DeepONet obtained from EKI can be efficiently computed as follows:
\begin{align}
    \overline{\mathcal{G}}_{\theta}(u^*)(y^*)&=\frac{1}{J}\sum_{j=1}^J\mathcal{G}_{\theta_I^{(j)}}({u}^*)(y^*),\label{al:mean}\\
    \text{std}(\mathcal{G}_{\theta}(u^*)(y^*))&=\sqrt{\frac{1}{J-1}\sum_{j=1}^J\left(\mathcal{G}_{\theta_I^{(j)}}({u}^*)(y^*)-\overline{\mathcal{G}}_{\theta}(u^*)(y^*)\right)^2}.\label{al:std}
\end{align}
These quantities provide important information about the ensemble and can be used to assess the quality and confidence of the model's predictions.

\subsection{Estimation of the observation error covariance matrix Q}
\label{subsec:Q}

The covariance matrix of the artificial dynamics $Q\in\mathbb{R}^{N_\theta\times N_\theta}$ in  \eqref{al:EKI-2-1} is a hyperparameter of the EKI, which helps prevent the collapse of the ensemble \cite{huang2022iterated}. The choice of this hyperparameter is important for ensuring the ensemble does not collapse; however, there is no obvious choice a priori of this matrix.

For this paper, we assume the covariance has the following form: $Q=\omega^2 I_{N_\theta}$, and shall attempt to learn a reasonable choice of the parameter $\omega>0$. We reserve a subset of input/output function pairs (denoted $\mathcal{D}^q$) disjoint from the training input/output function pair samples.
At the $i$th EKI iteration, we sample $m_q$ datapoints from $\mathcal{D}^q$ without replacement to obtain subset $\mathcal{D}^q_i$. We define the corresponding measurement operator $\mathcal{H}^q_i$, defined similarly to \eqref{al:H_op} over this subset. Denote the mean and standard deviation of this measurement operator applied to the ensemble as follows:

\begin{align}
\text{mean}(\mathcal{H}^q_i(\theta_i))&= \frac{1}{J}\sum_{j=1}^{J} \mathcal{H}^q_i(\theta_i^{(j)}), \\
    \text{std}(\mathcal{H}^q_i(\theta_i)) &=\sqrt{\frac{1}{J-1}\sum_{j=1}^{J} (\mathcal{H}^q_i(\theta_i^{(j)}) - \text{mean}(\mathcal{H}^q_i(\theta_i)) \odot (\mathcal{H}^q_i(\theta_i^{(j)}) - \text{mean}(\mathcal{H}^q_i(\theta_i))},
\end{align}
where $\odot$ is the elementwise product.
Based on these notations, we introduced a quantity indicating if the magnitude of the standard deviation $||\text{std}(\mathcal{H}^q_i(\theta_i))||_2$ is larger than the difference between the groundtruth data and the predicted ensemble mean $||\mathcal{D}^q_i -\text{mean}(\mathcal{H}^q_i(\theta_i))||_2$:
\begin{align}
f_{i} &= \frac{||\text{std}(\mathcal{H}^q_i(\theta_i))||_2 - ||\mathcal{D}^q_i -\text{mean}(\mathcal{H}^q_i(\theta_i))||_2}{||\mathcal{D}^q_i||_2}.\label{al:metric_v}
\end{align}
Based on these notions, we define a window of $W_q$ iterations, a threshold  $\tau>0$, and a step size $\alpha\in(0,1)$. Given $Q_i=\omega_{i}^2I_{N_\theta}$ at $i$-th iteration, we define $Q_{i+1}=\omega_{i+1}^2I_{N_\theta}$, where $\omega_{i+1}$ will be updated as follows:
\begin{align}
\omega_{i+1}\equiv \begin{cases}(1+\alpha)\omega_{i},& \text{median}(f_{i},...,f_{i-W_q})<-\tau,\\
 (1-\alpha)\omega_{i},&\text{median}(f_{i},...,f_{i-W_q})\geq \tau,\\
 \omega_{i},&\text{otherwise}.
 \end{cases}\label{al:Q}
 \end{align}
This particular choice aims to ensure that the standard deviation provided by the DeepONet ensemble is around the same magnitude as the difference between the predictive mean of  DeepONet  and the validation data, which may serve as a proxy of UQ.

\begin{figure}
\centering
\includegraphics[scale=0.5]
{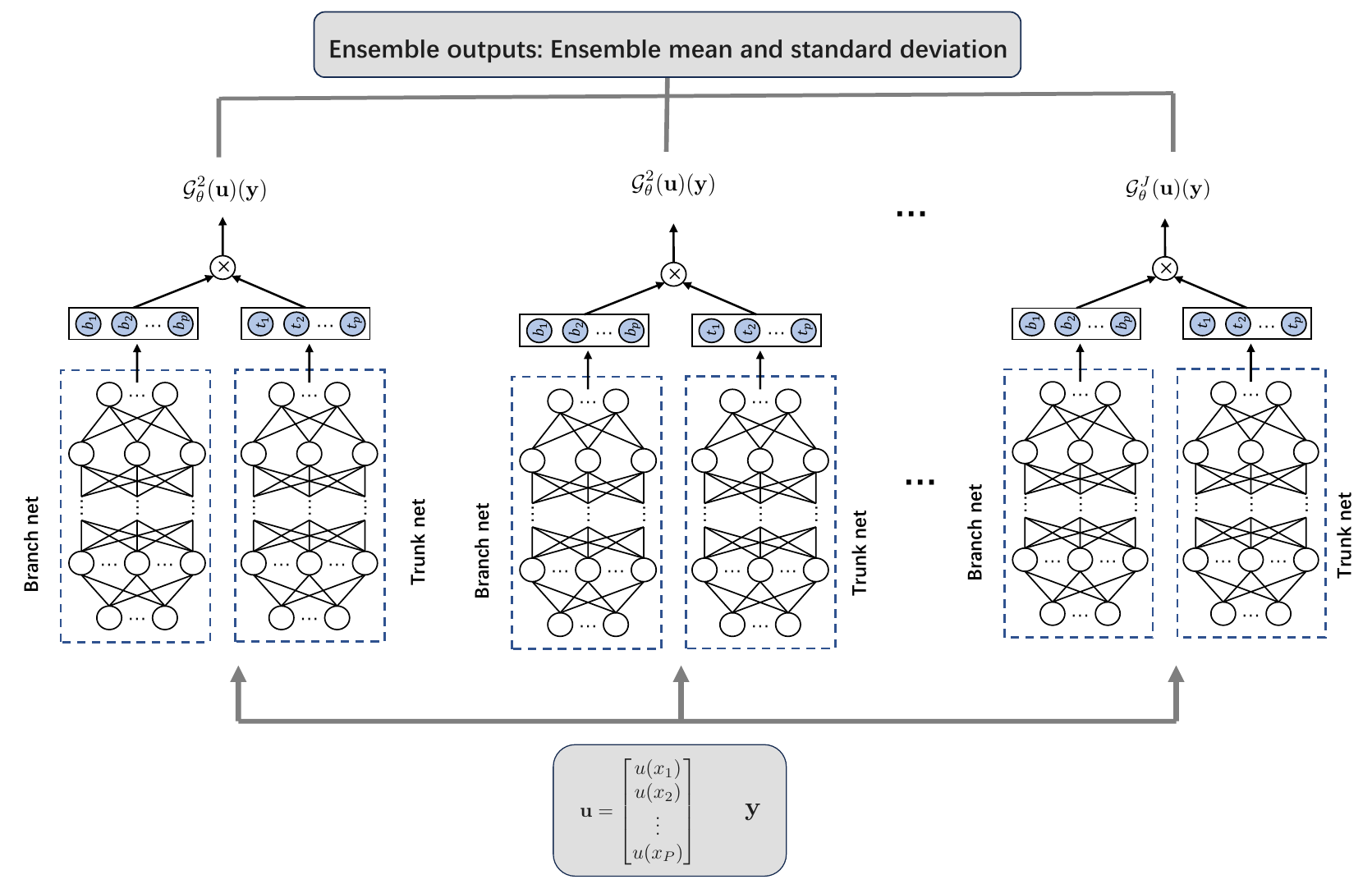}
\caption{Schematic of the EKI B-DeepONet architecture. Using the structure of a single DeepONet seen in Fig. \ref{fig:nn}, the ensemble consists of $J$ DeepONets with different sets of neural network parameters.}
\label{fig:nn_ensemble}
\end{figure}

\subsection{Stopping Criterion}
\label{subsec:sc}
Most EKI approaches utilize discrepancy-based stopping criteria \cite{Iglesias_2013} where the discrepancy refers to the weighted difference between the observed data and the predicted observation data:
\begin{align}
\left(\mathbf{y}-\frac{1}{J}\sum_{j=1}^J \mathcal{H}\left( \theta^{(j)}_i\right)\right)^TR^{-1}\left(\mathbf{y}-\frac{1}{J}\sum_{j=1}^J\mathcal{H}\left(\theta^{(j)}_i\right)\right)<\tau,
\end{align}
where $\tau>0$ is a threshold parameter. 

In \cite{Pensoneault2023}, the authors employed a discrepancy-based stopping criterion for early termination of the EKI method. This approach was effective in a setting that utilized the full batch measurement operator. However, as we employ a mini-batch approach in this paper, the convergence of this term becomes noisy, diminishing the effectiveness of the approach described in the aforementioned paper. Similar to Section \ref{subsec:Q}, we reserve a subset of the input/output function pairs not used to train the DeepONet, $D^s$, to terminate the training process. We further define $R^s$ as the covariance matrix corresponding to this subset $\mathcal{D}^s$, defined similar to \eqref{al:R_def}.

On the $i$th EKI iteration, similar to in \ref{subsec:Q}, we subsample $m_s$ samples from $D^s$ without replacement. We denote this subset $\mathcal{D}^s_i$, define the corresponding measurement operator $\mathcal{H}^s_i$, and the sub-matrix $R^s_i$ corresponding to the covariance matrix of this dataset.

To alleviate the impact of noise on early stopping, at $i$th iteration, we introduce a smoothed discrepancy metric over a window of $W$ iterations as follows:
\begin{align}
\tilde{D}_i &= \frac{1}{W}\sum_{w=1}^{W}({\mathcal{D}^s_{i-w}}-\text{mean}(\mathcal{H}^s_{i-w}(\theta_{i-w}))^T(R^s_{i-w})^{-1}({\mathcal{D}^s_{i-w}}-\text{mean}(\mathcal{H}^s_{i-w}(\theta_{i-w})),
\end{align}
where  
\begin{align}
\text{mean}(\mathcal{H}^s_k(\theta_k))&= \frac{1}{J}\sum_{j=1}^{J} \mathcal{H}^s_k(\theta_k^{(j)})
\end{align}  
We terminate the EKI iterations if the minimum smoothed discrepancy $\tilde{D}_i$ does not decrease over $K$ successive iterations. Based on our test, this smoothing technique helps reduce fluctuations in the discrepancy caused by mini-batching and facilitates a more effective early stopping criterion compared to the stopping criterion proposed in \cite{Pensoneault2023}. We remark that this idea bears similarities to some of the early stopping criteria for neural networks discussed in \cite{Prechelt2012}. 
Figures\section{Numerical Examples}
\label{sec:NumExample}
In this section, we test the performance the proposed EKI B-DeepONet inference algorithm using three examples: an anti-derivative operator, a gravity pendulum, and a diffusion-reaction system. 

\subsection*{Data Generation} For each example, we generate $1000$ input functions and compute the corresponding output functions using the associated numerical solvers in \cite{yang2022scalable}. Of the $1000$ pairs, $800$ input/output pairs are utilized in the training of the DeepONet, $100$ input/output pairs are utilized for learning $Q$, and $100$ input/output pairs are utilized for the stopping criterion. Additionally, $1000$ input/output function pairs are generated in the same way for testing.

{\bf NN design and training}. The DeepONets within these experiments utilize fully connected neural networks with 3 hidden layers with 128 hidden units per layer, 128 output units, and ReLU activation function for the branch net $b(u)$ and the Tanh activation function for the trunk net $t(y)$. The total number of parameters of the network for each example needs to be inferred listed in Table. \ref{tab:param_num}. We initialize each ensemble member using a standard Gaussian prior for the neural network parameters $\theta\sim\mathcal{N}(0,I_{N_\theta})$. Finally, we choose the standard deviation $\sigma_l$ of the likelihood in \eqref{al:likelihood2} to be $1\%$ or $5\%$ of the maximum of the output function over the output domain depending on the noise added to the dataset. 

To train the DeepONets, from the total set of training data, at each EKI iteration, we draw a total of $m_t=500$ sample output data from the total set of data without replacement. Similarly, for the leaning of $Q$ and the stopping criterion mentioned in Sec. \ref{subsec:Q} and Sec. \ref{subsec:EKI-B-DO}, we select randomly selecting $m_q=500$ and $m_s=500$ samples without replacement respectively from the corresponding disjoint data sets.

To evaluate the robustness of the method, we corrupt these samples by adding zero mean Gaussian noise with standard deviation corresponding to $1\%$ and $5\%$ of the maximum of the output function over the output domain.

{\bf EKI setup.} For EKI, we utilize $J=5000$ ensemble members. We shall adopt the learning $Q$ approach for all examples and initialize the scale parameter of the standard deviation of the artificial dynamics from \eqref{al:Q} to be $\omega_{0}=0.01$ and the corresponding $\alpha=0.05$. One simple example to illustrate the benefits of learning $Q$ approach is presented in Appendix A.  Additionally, we set the stopping criterion hyperparameters $W=10$ and $K=100$ in section \ref{subsec:sc}. 

\begin{table}[ht]
\centering
\begin{tabular}{|l|l|}
\hline
& $\boldsymbol{\theta}$ size \\ \hline
Example \ref{subsec:ex1}  & 79232  \\ \hline
Examples \ref{subsec:ex3}, \ref{subsec:ex2} & 79360 \\
\hline
\end{tabular}
\caption{The number of neural network parameters $\boldsymbol{\theta}$ for each example. \label{tab:param_num}}
\end{table}

\subsection*{Performance Metrics}
We shall examine the performance of the EKI B-DeepONets 
over a set of $N_t=1000$ testing input/output function pairs $\{(u^{l}_t, s^{l}_t)\}_{l=1}^{N_t}$,  denoted the ``testing" data. We define the the $l$-th output function $s^{l}_t$ at query locations $\{y_k^l\}$ as $\mathcal{D}^l_t = \{\mathcal{G}(u^{l}_t)(y_{k}^l)\}_{k=1}^{P}$. Similarly,   $\tilde{\mathcal{D}}^l_t$ is defined for the corresponding predicted outputs by DeepONets. 

Specifically, we examine the performance of the EKI B-DeepONets based on three metrics, the mean relative error, the uncertainty, and the coverage of the two standard deviation confidence interval over the test set. For the $l$-th input and output pair (with $P$ query locations) in the test set, we first define: 
\begin{itemize}
    \item The relative error,
    \begin{equation}
e^{l}_{t} = \frac{||\mathcal{D}^l_t -\text{mean}({\tilde{\mathcal{D}}^l_t})||_2}{||{\mathcal{D}^l_t}||_2},
    \end{equation}
    measures how closely the predicted mean output data matches the true output data. 
    \item  The uncertainty metric,
        \begin{equation}
    q^{l}_{t} =\frac{||\text{std}(\tilde{\mathcal{D}}^l_t)||_2}{||\mathcal{D}^l_t||_2},
        \end{equation}
   is computed as the average standard deviation of the predictions, scaled by the magnitude of the true output functions, measuring the variability in the predicted output functions.  
    \item  The confidence interval coverage metric,
       \begin{equation}
    c^{l}_{t} = \frac{\#\text{ }\mathcal{D}^l_t\text{ contained in }\text{mean}({\tilde{\mathcal{D}}^l_t})\pm2\text{ std}(\tilde{\mathcal{D}}^l_t)}{P},
       \end{equation}
   assesses the model's ability to provide meaningful uncertainty estimates that accurately capture the true values. It measures the percentage of test samples contained within  two standard deviation confidence interval of the predicted sample. 
\end{itemize}

We additionally define the mean of the above metrics across the entire testing set as follows:
\begin{align}
q_{t} &=\frac{1}{N_t}\sum_{l=1}^{N_t} q^{l}_{t}, \quad
e_{t} = \frac{1}{N_t}\sum_{l=1}^{N_t} e^{l}_{t},\quad
c_{t} = \frac{1}{N_t}\sum_{l=1}^{N_t} c^{l}_{t}.\label{al:metric_t_sum}
\end{align}

All tests are carried out using the JAX library on a NVIDIA A10 graphics card with 24GB of RAM. The Ensemble DeepONet code is modified based on the UQ-DeepONet \cite{yang2022scalable}. For each example, we also report the walltime of the training iterations. Walltime and number of training iteration are presented as the average of 10 independent runs.

\subsection{Anti-derivative}
\label{subsec:ex1}
We first consider the problem of learning the anti-derivative operator \cite{yang2022scalable} as follows:
\begin{align}
\frac{ds}{dx} &= u(x),\quad x\in[0,1],\\
s(0) &= 0,
\end{align}
where $u(x)$ is the input function and $s(x)$ is the corresponding output function. Input functions $u$ are generated from Gaussian process $u\sim\mathcal{GP}(\mu,\mathcal{K})$ 
\begin{align}
    \mu(x)&\equiv 0\label{al:gp_1},\\
    \mathcal{K}(x,x')&=\alpha^2\exp\left(-\frac{(x-x')^2}{2\ell^2}\right) \label{al:gp_2},
\end{align}
where $\alpha$ is the scale parameter and $\ell$ is the length scale  with the corresponding input sensors $u$ at $m=100$ equally spaced locations along $x\in [0,1]$. For this example, $\alpha=1.0$ and $\ell=0.2$. Training output data is generated at 100 equally spaced locations along $x \in [0,1]$ for each corresponding output function $s$.

\subsubsection{Small Noise}
We first consider training the DeepONets with training output data $\{s^l\}$ corrupted by Gaussian noise, with zero mean and a standard deviation set to  $1\%$ of the maximum absolute value of the corresponding output function $s^l$.

\begin{figure}[ht]
\centering
\subfloat[]{\includegraphics[width=.5\textwidth]{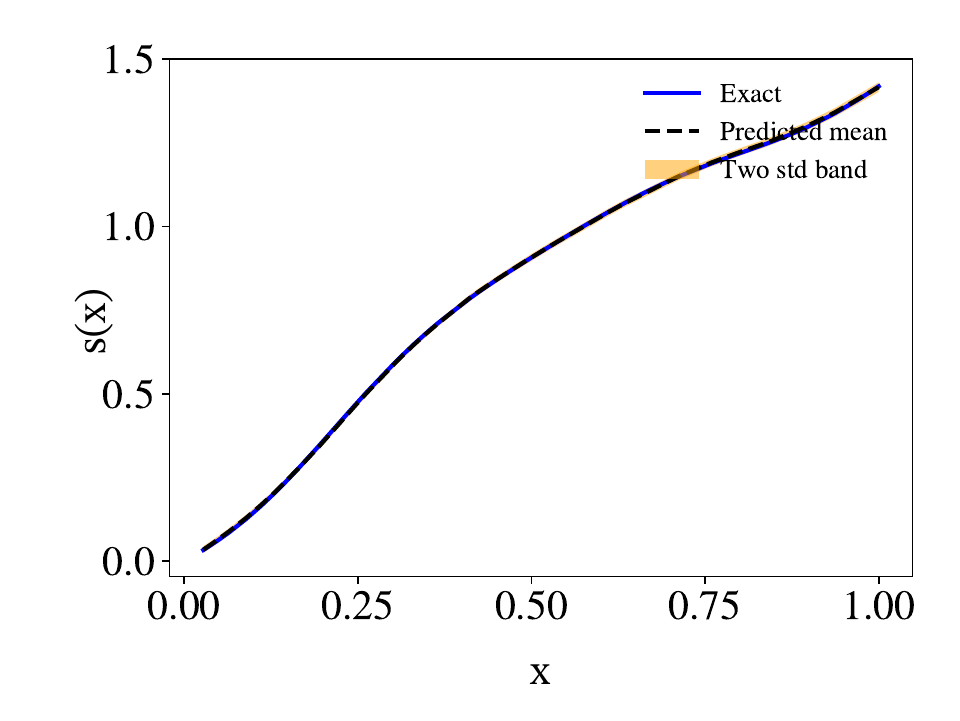}}
\hfill
\subfloat[]{\includegraphics[width=.5\textwidth]{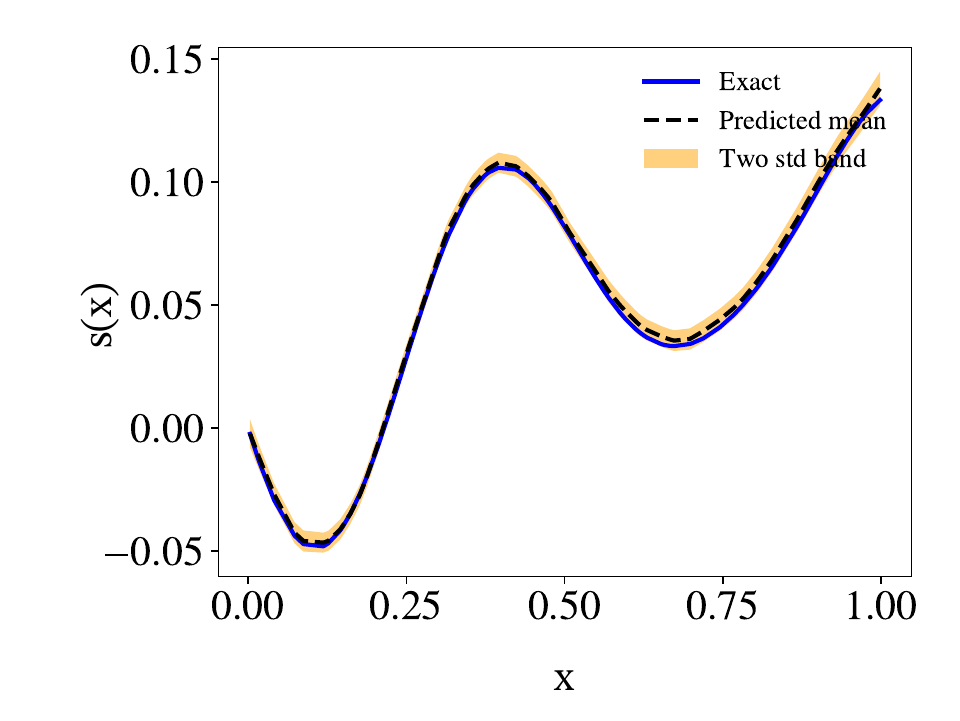}}\\
\subfloat[]{\includegraphics[width=.5\textwidth]{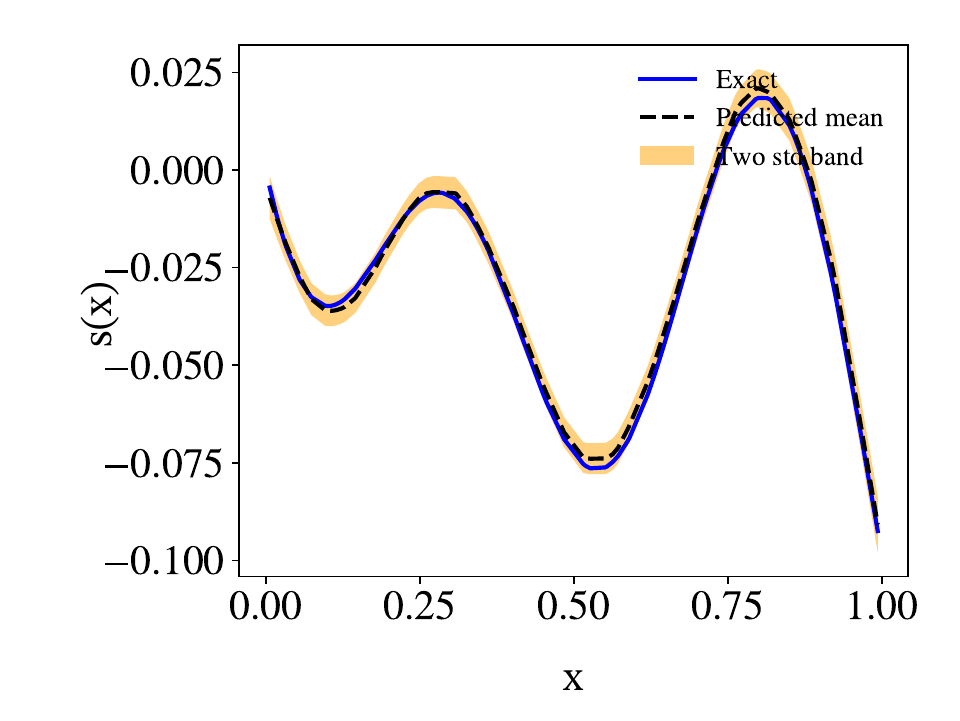}}
\caption{Example \ref{subsec:ex1} small noise: (a) and (b) two sample output function approximations with two standard deviation confidence intervals; (c) worst case scenario approximation of the output samples.}
\label{fig:ex_1_samples_small}
\end{figure}
Figure \ref{fig:ex_1_samples_small} shows several sample solutions from the testing data set and the corresponding estimated mean and two standard deviation confidence intervals by DeepONet. The ensemble mean well approximates the samples across several orders of magnitudes, including in the worst-case sample. Moreover, it is worth noting that the confidence interval associated with these samples can effectively capture a significant portion of the ground truth samples.

\begin{figure}[ht]
\centering
\includegraphics[scale=0.45]{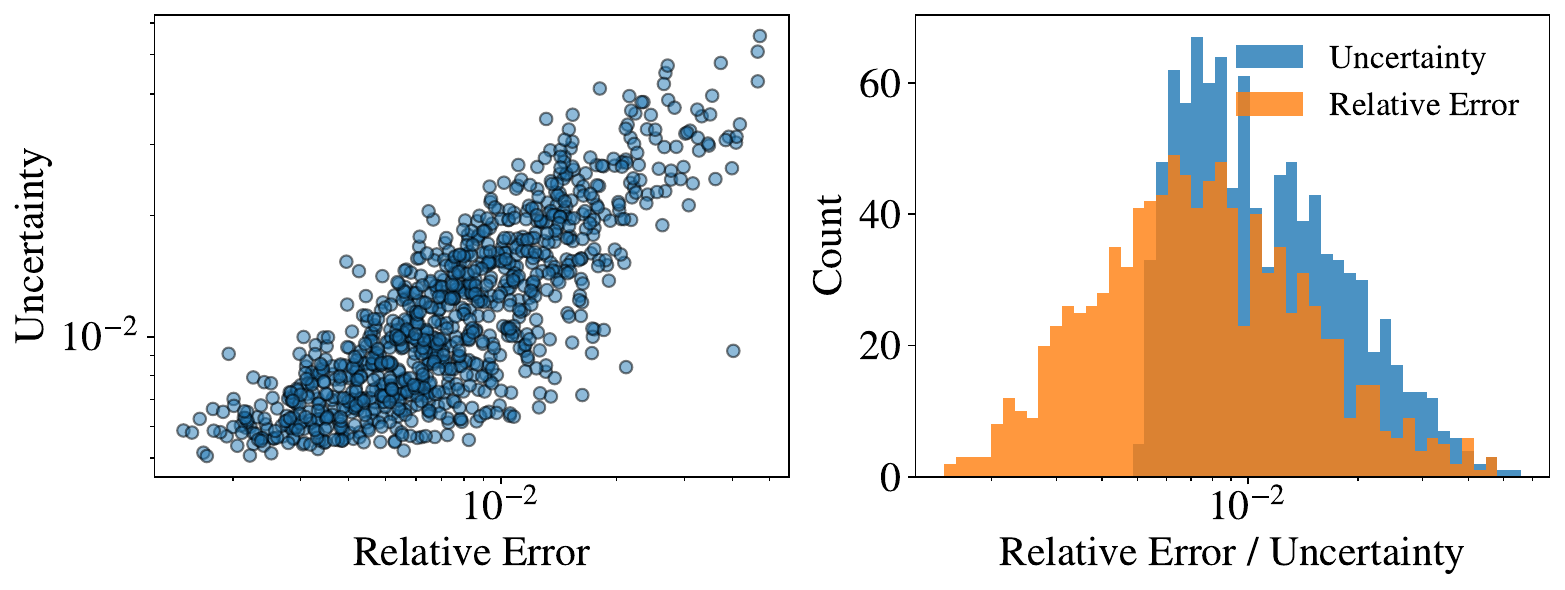}
\caption{Example \ref{subsec:ex1} small noise: \emph{(Left)}  Uncertainty versus the relative error for outputs from the EKI DeepONet. \emph{(Right)} Estimates of marginal densities of the relative error and uncertainties of the testing sample outputs.}
\label{fig:Ex_1_uq_small}
\end{figure}

Figure \ref{fig:Ex_1_uq_small} (left) presents the mean relative error versus the uncertainty of the test sample output. A clear positive correlation between the relative error and uncertainty can be observed, indicating a high uncertainty serves as an effective indicator of a larger error in the corresponding mean prediction. The corresponding marginal densities are also presented in Figure \ref{fig:Ex_1_uq_small} (right) for both the relative error and uncertainty. Notably,  the distribution of the uncertainty by EKI-DeepONets provides a conservative approximation of the corresponding relative error distribution.

\begin{table}[ht]
\centering
\begin{tabular}{|l|l|l|l|l|}
\hline
Relative Error & Uncertainty & Coverage & Training Time & Iterations \\ \hline
  0.009 & 0.014 &  0.985 & 239.4 seconds & 814 \\ \hline
\end{tabular}
\caption{Example \ref{subsec:ex1} small noise: mean relative error, uncertainty, coverage metrics over the test dataset,
 training time, and the number of EKI iterations. \label{tab:Ex_1_tab_small}}
\end{table}

Table \ref{tab:Ex_1_tab_small} shows the mean relative error, uncertainty, and coverage by EKI B-DeepONets over the test set. The mean relative error shows a $0.9\%$ error in the mean prediction and a corresponding $1.4\%$ uncertainty. Additionally, on average, $98.5\%$ of each truth sample fall within the two standard deviation confidence interval, indicating the provided confidence interval well  captures the mean error at the majority of the locations in the domain. Note that even though the number of parameter estimated is  79232, the training time is roughly $240$ s for 814 EKI iterations.

\subsubsection{Large Noise}
We now assume that the training output data $\{s^l\}$ is contaminated by independent Gaussian noise with zero mean. The standard deviation of the noise is set to $5\%$ of the maximum absolute value of the output function $s^l$ during the training of the DeepONets. 

\begin{figure}[ht]
\centering
\subfloat[]{\includegraphics[width=.5\textwidth]{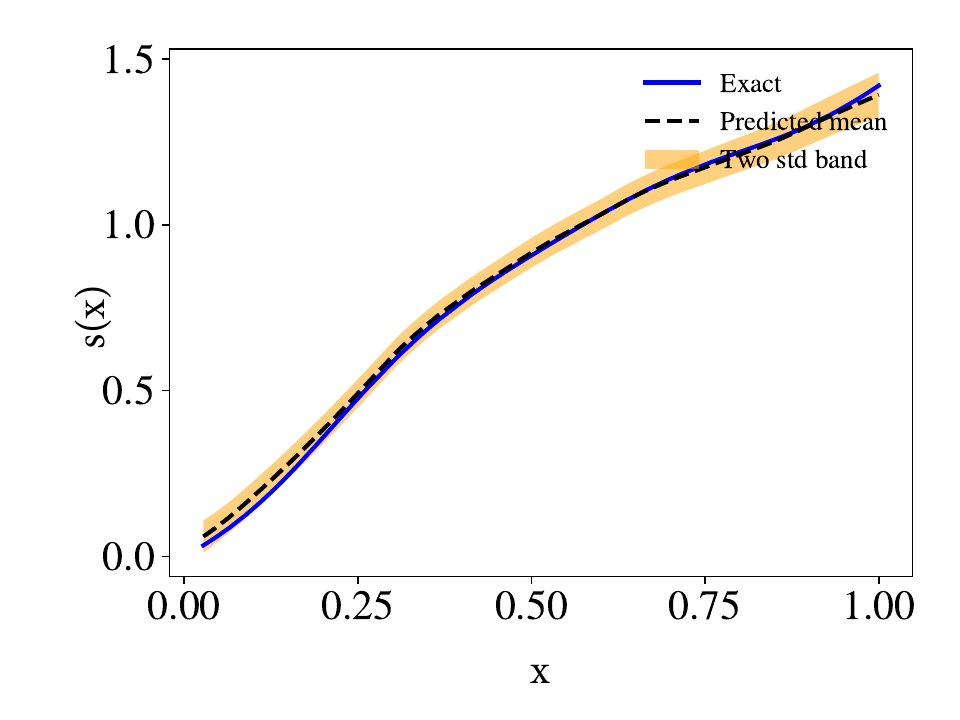}}
\hfill
\subfloat[]{\includegraphics[width=.5\textwidth]{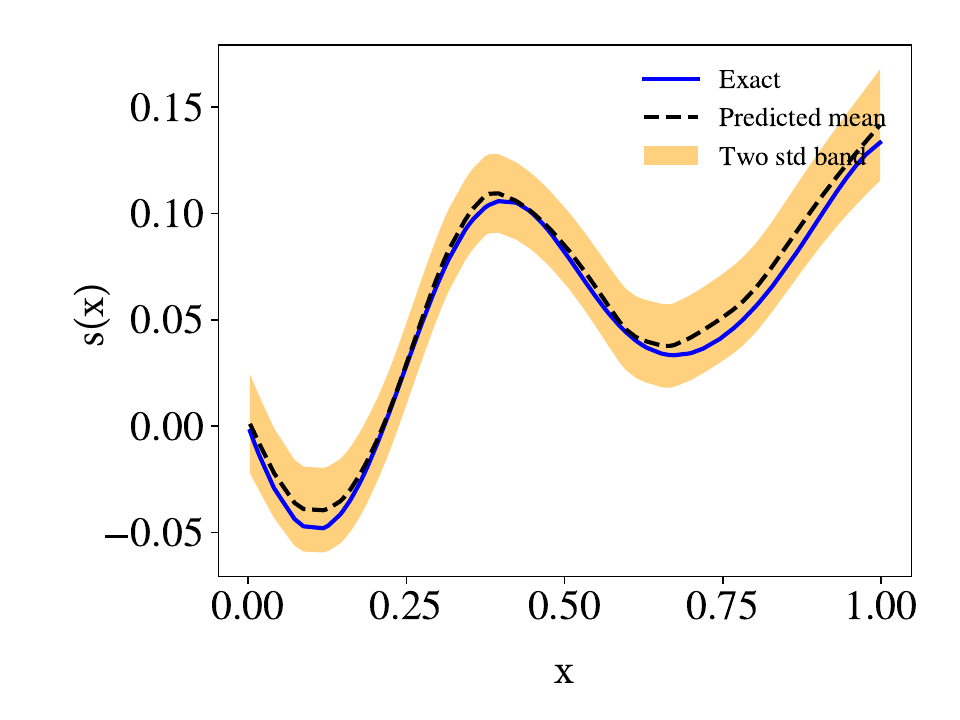}}\\
\subfloat[]{\includegraphics[width=.5\textwidth]{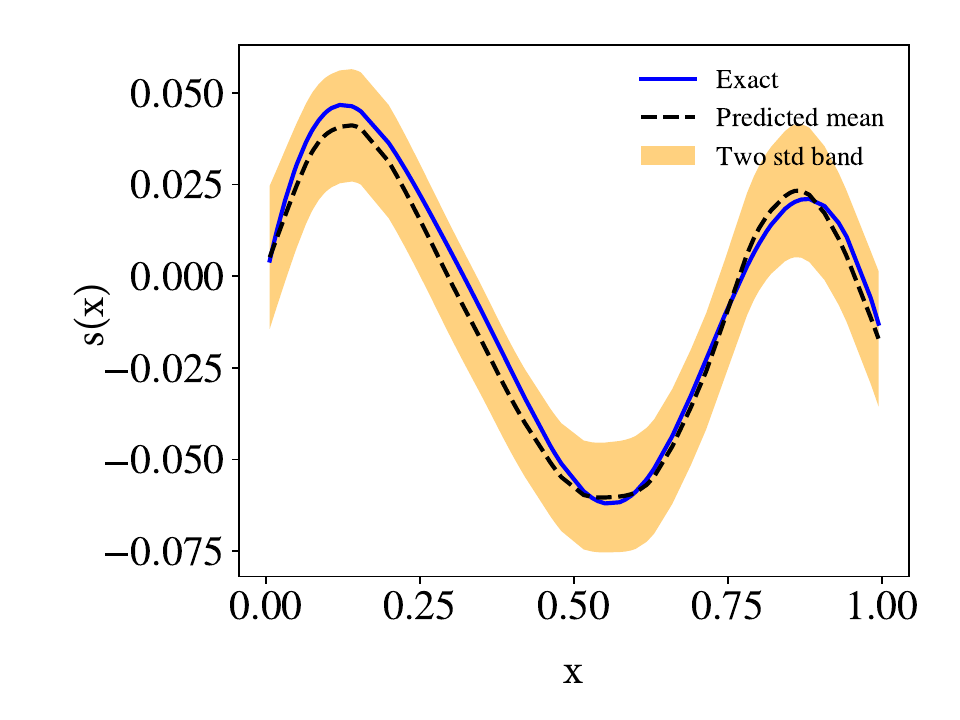}}
\caption{Example \ref{subsec:ex1} large noise: (a) and (b) two sample output approximations with two standard deviation confidence intervals; (c) worst case scenario approximation of the output samples.}
\label{fig:ex_1_samples_large}
\end{figure}
Figure \ref{fig:ex_1_samples_large} shows two randomly selected sample solutions and the worst-case scenario from the testing data set with the corresponding mean and two standard deviation confidence intervals by the DeepONets. In both cases, the ensemble means well approximate the truth. Notably, the predictions by DeepOnets show more conservative uncertainty than scenarios with smaller noise levels for all samples. Notably,  the worst-case scenario shows increased error compared to the small noise case but with a more conservative confidence interval. 

\begin{figure}[ht]
\centering
\includegraphics[scale=0.45]{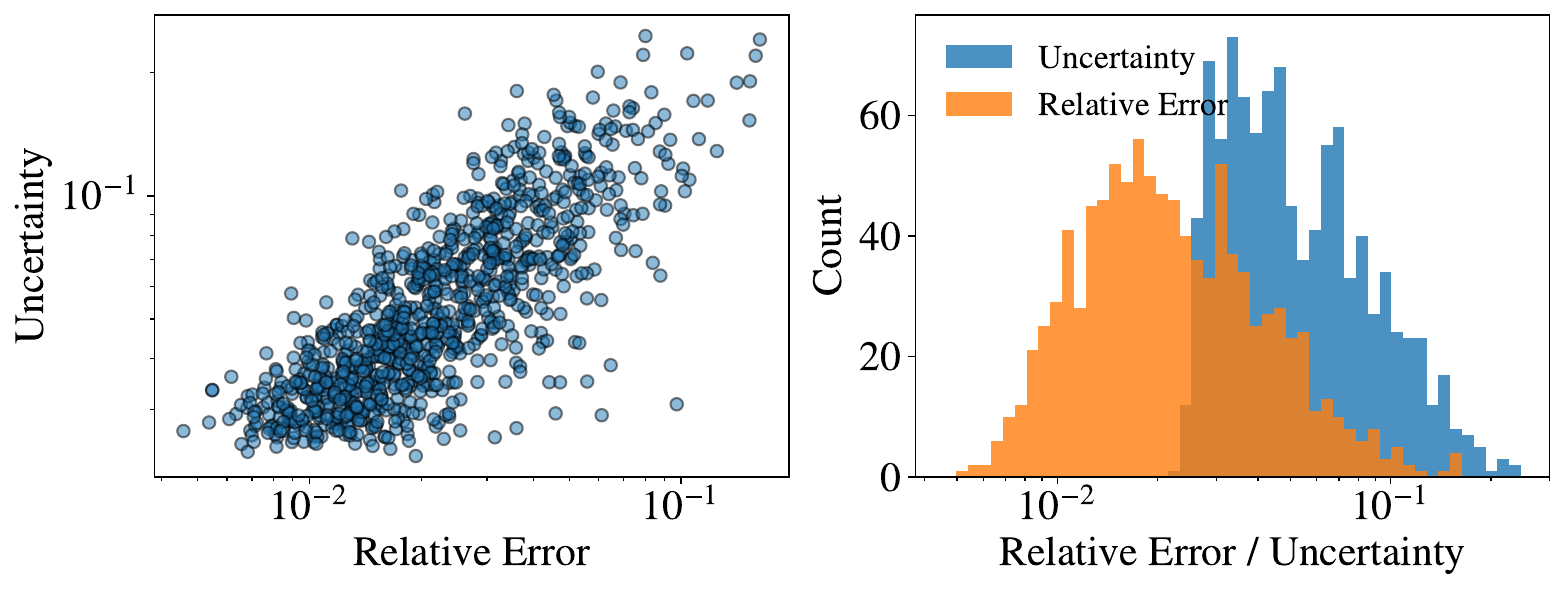}
\caption{Example \ref{subsec:ex1} large noise: \emph{(Left)}  Uncertainty versus relative error for outputs from the EKI DeepONet. \emph{(Right)} Histograms of relative error and uncertainties of the testing sample  outputs.}
\label{fig:Ex_1_uq_large}
\end{figure}

Figure \ref{fig:Ex_1_uq_large}  (left)  shows again a strong positive correlation between the relative error and uncertainty. While the relative error is larger for the outlier predictions, those points also exhibit higher uncertainty, indicating the uncertainty can serve as an informative error indicator in identifying outlier datasets. Furthermore, the corresponding marginal density of uncertainty  in Figure \ref{fig:Ex_1_uq_large}  (right)  appears to be a more conservative indicator for the relative error, especially when compared to the scenario with lower noise levels. Besides, the training time is roughly $230$ s for 615 EKI iterations.

\begin{table}[ht]
\centering
\begin{tabular}{|l|l|l|l|l|}
\hline
Relative Error & Uncertainty & Coverage & Training Time & Iterations\\ \hline
  0.027 & 0.070 &  0.997 & 229.8 Seconds & 615 \\ \hline
\end{tabular}
\caption{Example \ref{subsec:ex1} large noise: mean relative error, uncertainty, coverage metrics over the test dataset, training time, and the number of EKI iterations.  \label{tab:Ex_1_tab_large}}
\end{table}

Finally, Table \ref{tab:Ex_1_tab_small} shows a mean relative error of $2.7\%$ and uncertainty $7.0\%$, signifying a conservative average estimate across the test set. In this case, the larger conservative estimate corresponds to an average coverage of $99.7\%$ for each sample within the test set, highlighting the conservative nature of this estimate. 

\subsection{Reaction-Diffusion Equation}
\label{subsec:ex2}

Next, we consider 
the following 1D the Reaction-Diffusion Equation \cite{yang2022scalable}:
\begin{align}
    \frac{\partial s}{\partial t} &= \nu\frac{\partial^2 s}{\partial x^2} + ks^2 + u(x),\quad (x,t)\in[0,1]\times[0,1],\\
    s(x,0) &= 0.0,\quad x\in[0,1],\\
    s(0,t) &= 0.0,\quad x\in[0,1],\\
    s(1,t) &= 0.0,\quad t\in[0,1].
\end{align}
where $k=0.01$, $\nu=0.01$, $u$ is the input function and $s$ is the corresponding output function. The input functions $u$ are similarly generated by the Gaussian process as in \eqref{al:gp_1}-\eqref{al:gp_2}. The input sensors $u$ are placed along the spatial domain with $m=100$ equally spaced input locations. Training output data is generated at $100\times100$  equally spaced grid locations in the space and time domain $[0,1]\times[0,1]$  for each corresponding output function $s$. 

\subsubsection{Small Noise}
We first examine a small-noise scenario where the training output data is corrupted by zero-mean Gaussian noise. The standard deviation of this noise is set to $1\%$ of the maximum absolute value of each output function $s^l$.

\begin{figure}[ht]
\centering
\subfloat[]{\includegraphics[width=1\textwidth]{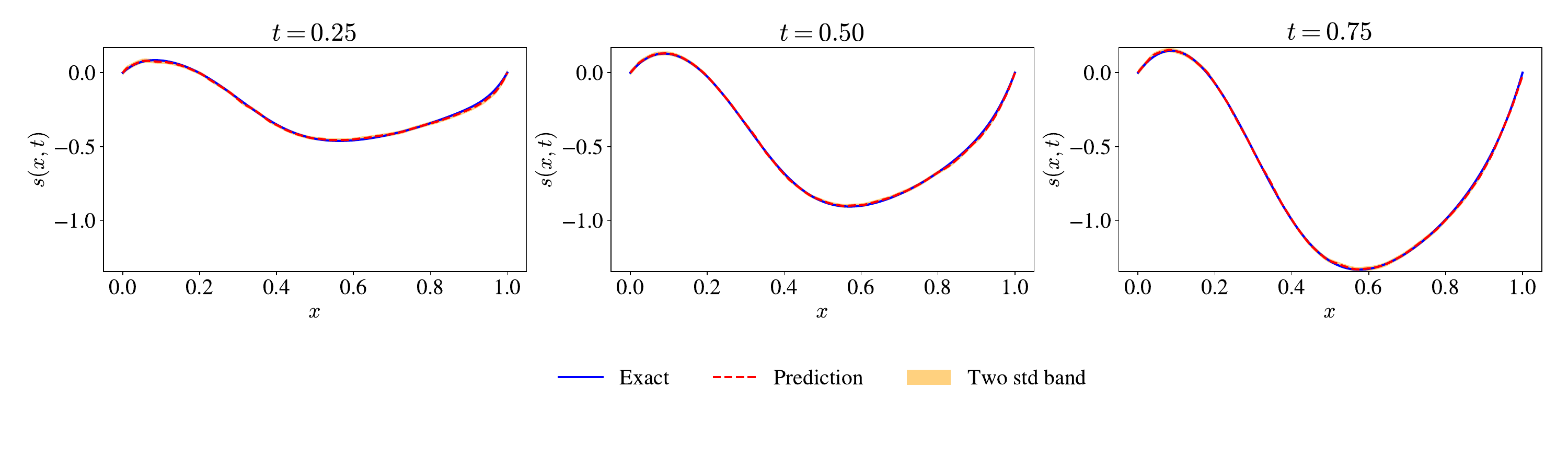}}
\\
\subfloat[]{\includegraphics[width=1\textwidth]{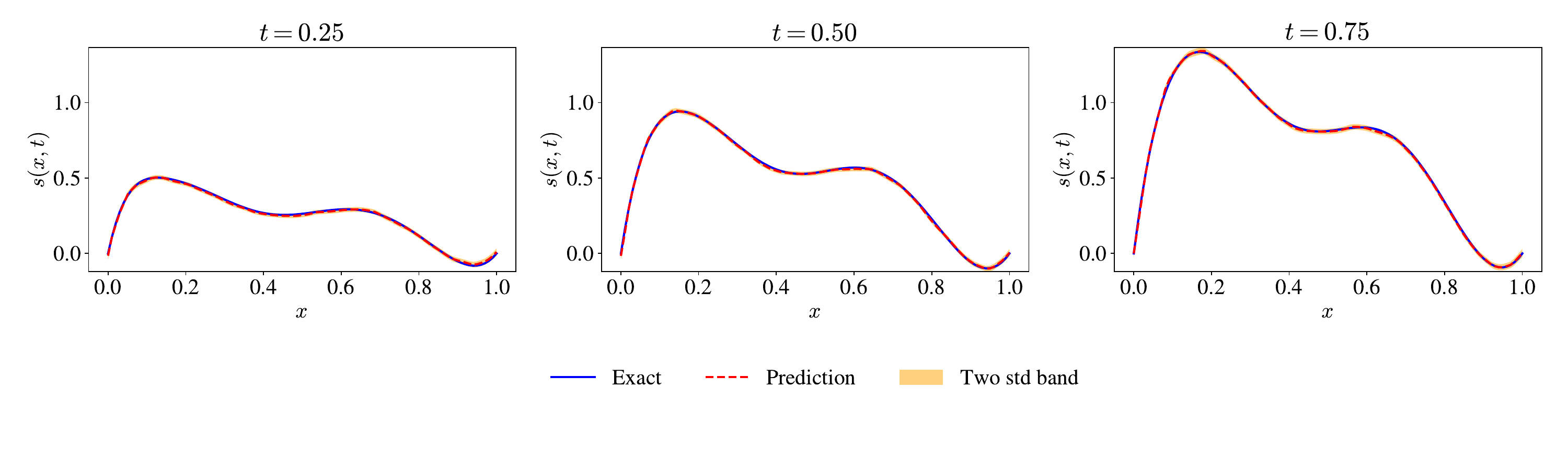}}\\
\subfloat[]{\includegraphics[width=1\textwidth]{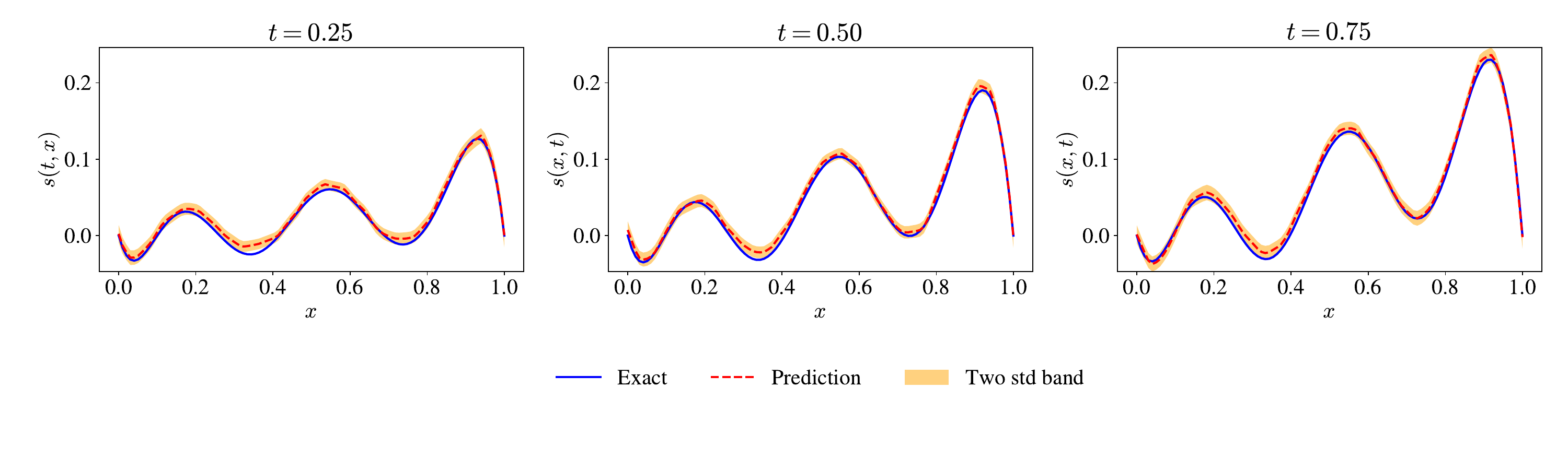}}
\caption{Example \ref{subsec:ex2} small noise: (a) and (b) two sample output function approximations with two standard deviation confidence intervals; (c) worst case scenario approximation of the output samples.}
\label{fig:ex_2_samples_small}
\end{figure}

Figure \ref{fig:ex_2_samples_small}  shows two randomly selected sample solutions and the worst-case scenario from the testing data set with the corresponding  posterior mean and two standard deviation confidence intervals by DeepONets at several different snapshots. 
It is evident that the posterior mean of the DeepONets well approximates the solution. Additionally, the corresponding approximation has a high degree of certainty, as seen in the narrow confidence intervals.

\begin{table}[ht]
\centering
\begin{tabular}{|l|l|l|l|l|}
\hline
Relative Error & Uncertainty & Coverage & Training Time &Iterations \\ \hline
  0.016 & 0.030 &  0.993 & 481.8 Seconds & 1429  \\ \hline
\end{tabular}
\caption{Example \ref{subsec:ex2} small noise: mean relative error, uncertainty, coverage metrics over the test dataset, training
  time, and the number of EKI iterations. \label{tab:Ex_2_tab_small}}
\end{table}

Table \ref{tab:Ex_2_tab_small} shows the corresponding mean metrics across the testing dataset. In this case, the mean relative error and mean uncertainty are below $3\%$, indicating accurate predictions with a high level of confidence across the dataset. Additionally, the corresponding coverage metrics of around $99.3\%$ suggest that few outliers lie outside the two standard deviation confidence interval. 

\begin{figure}[ht]
\centering
\includegraphics[scale=0.45]{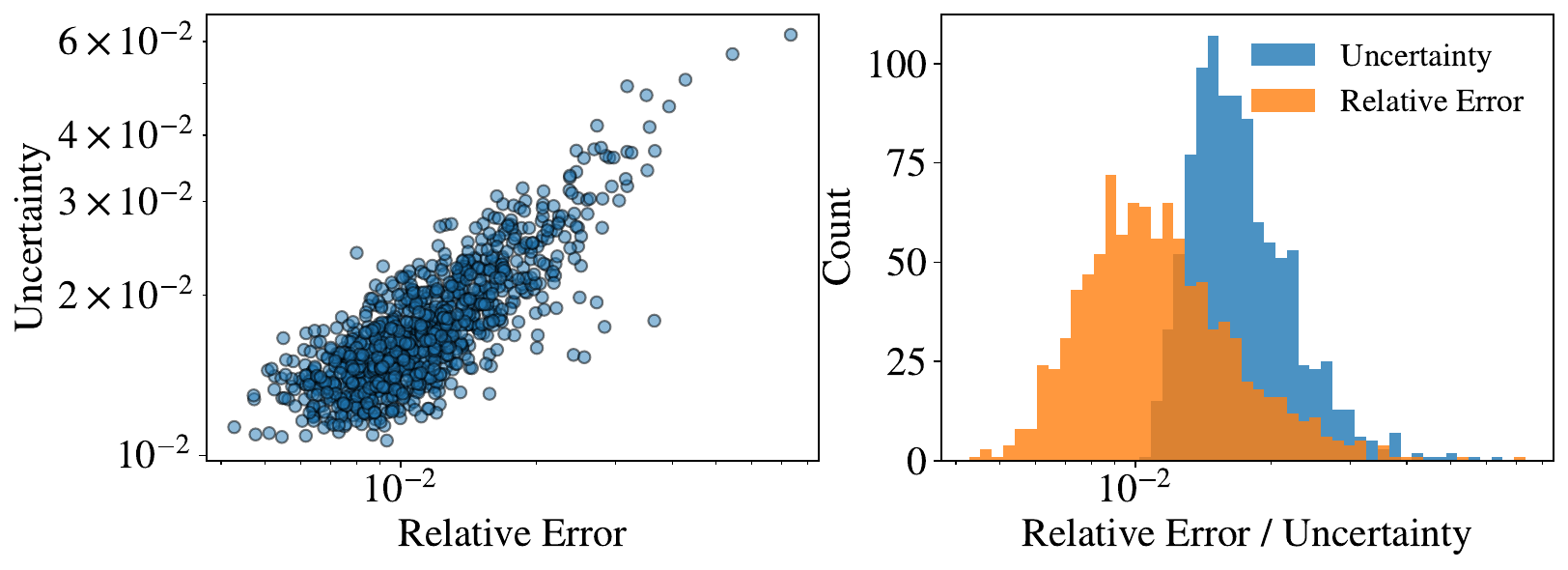}
\caption{Example \ref{subsec:ex2} small noise: \emph{(Left)}  Uncertainty versus relative error for outputs from the EKI DeepONet. \emph{(Right)} Histograms of relative error and uncertainties of the testing sample  outputs.}
\label{fig:Ex_2_uq_small}
\end{figure}

Figure \ref{fig:Ex_2_uq_small} (left) illustrates a strong correlation between uncertainty and relative error.
Notably, the samples with the larger relative errors tend to coincide with those exhibiting a higher level of uncertainty. Additionally, when considering the marginal distributions of relative error and uncertainty in Figure \ref{fig:Ex_2_uq_small} (right), it becomes evident that the uncertainty distribution provided by EKI-DeepONets offers a conservative estimate of the distribution of the corresponding relative error. Besides, the training time is roughly $480$ s for 1429 EKI iterations.

\subsubsection{Large Noise}
We now investigate the DeepONets trained with the output data corrupted by zero mean Gaussian noise with a standard deviation equal to  $5\%$ of the maximum absolute value of each output function $s^l$.

\begin{figure}[ht]
\centering
\subfloat[]{\includegraphics[width=1\textwidth]{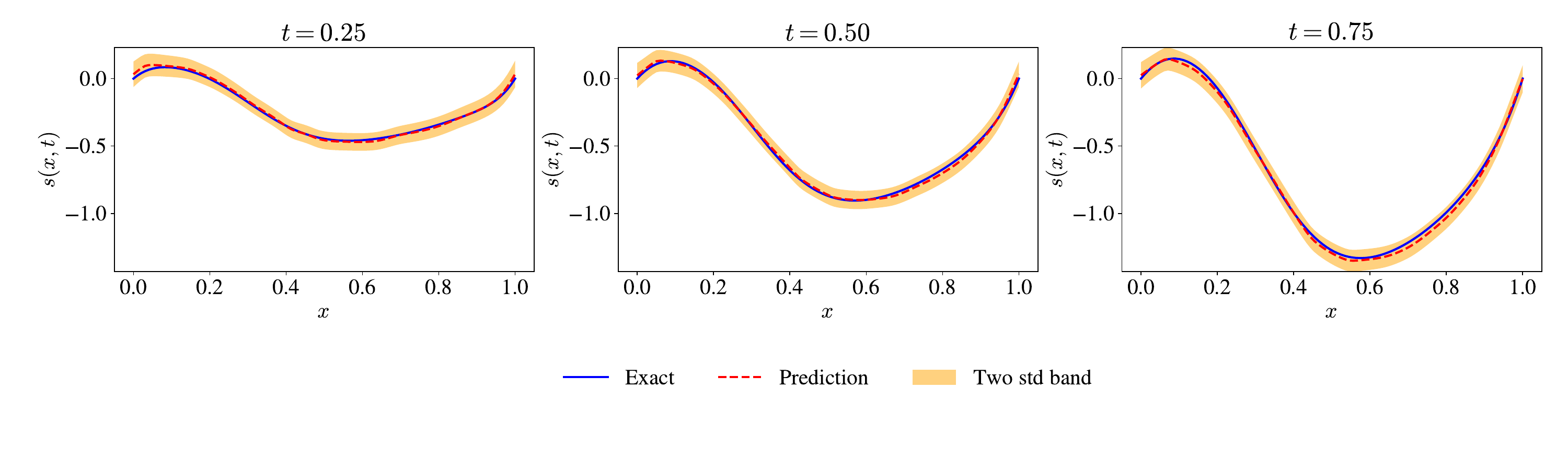}}
\\
\subfloat[]{\includegraphics[width=1\textwidth]{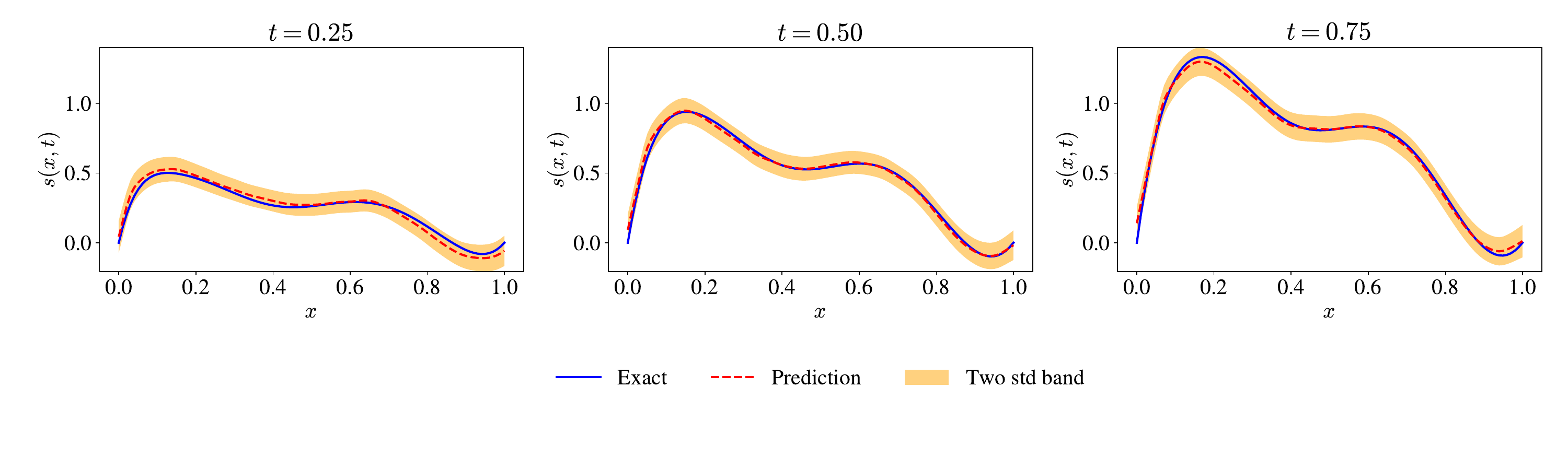}}\\
\subfloat[]{\includegraphics[width=1\textwidth]{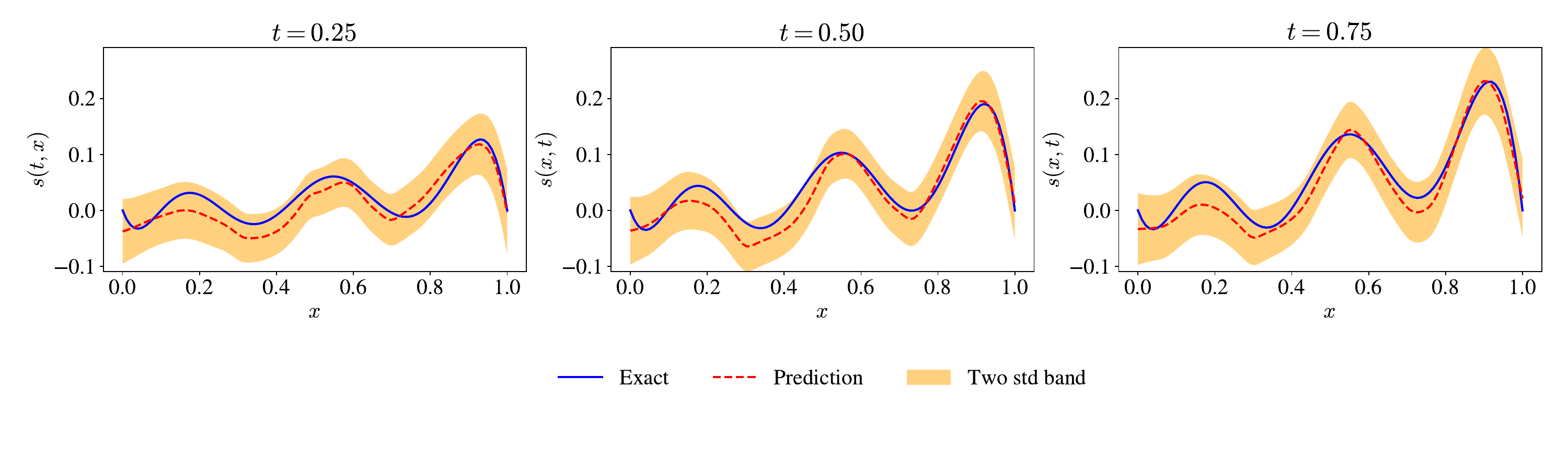}}
\caption{Example \ref{subsec:ex2} large noise: (a) and (b) two sample output function approximations with two standard deviation confidence intervals; (c) worst case scenario approximation of the output samples.}
\label{fig:ex_2_samples_large}
\end{figure}

Figure \ref{fig:ex_2_samples_large} shows the ensemble mean and two standard deviation confidence intervals of two samples and the worst case sample from the training dataset at different time steps, along with the corresponding truth solutions. As expected, the larger noise case has less accurate mean estimates than the $1\%$ noise case. Nonetheless, this is accompanied by larger confidence intervals, indicating higher uncertainty.

\begin{table}[ht]
\centering
\begin{tabular}{|l|l|l|l|l|}
\hline
Relative Error & Uncertainty & Coverage & Training Time & Iterations\\ \hline
  0.051 & 0.084 &  0.993 & 483.4 Seconds & 1554  \\ \hline
\end{tabular}
\caption{Example \ref{subsec:ex2} large noise: mean relative error, uncertainty, coverage metrics over the test dataset, training
  time, and the number of EKI iterations. }
 \label{tab:Ex_2_tab_large}
\end{table}

Table \ref{tab:Ex_2_tab_large} shows a larger relative error for the posterior mean and the corresponding uncertainty, $5.1\%$ and $8.4\%$, respectively.  As expected, these values are greater when compared to the scenario with lower noise levels.  Furthermore, the coverage of $99.3\%$ observed within the two-standard deviation confidence interval suggests that the uncertainty estimate tends to be conservative. Besides, the training time is roughly $483$ s for 1554 EKI iterations. 

\begin{figure}[ht]
\centering
\includegraphics[scale=0.45]{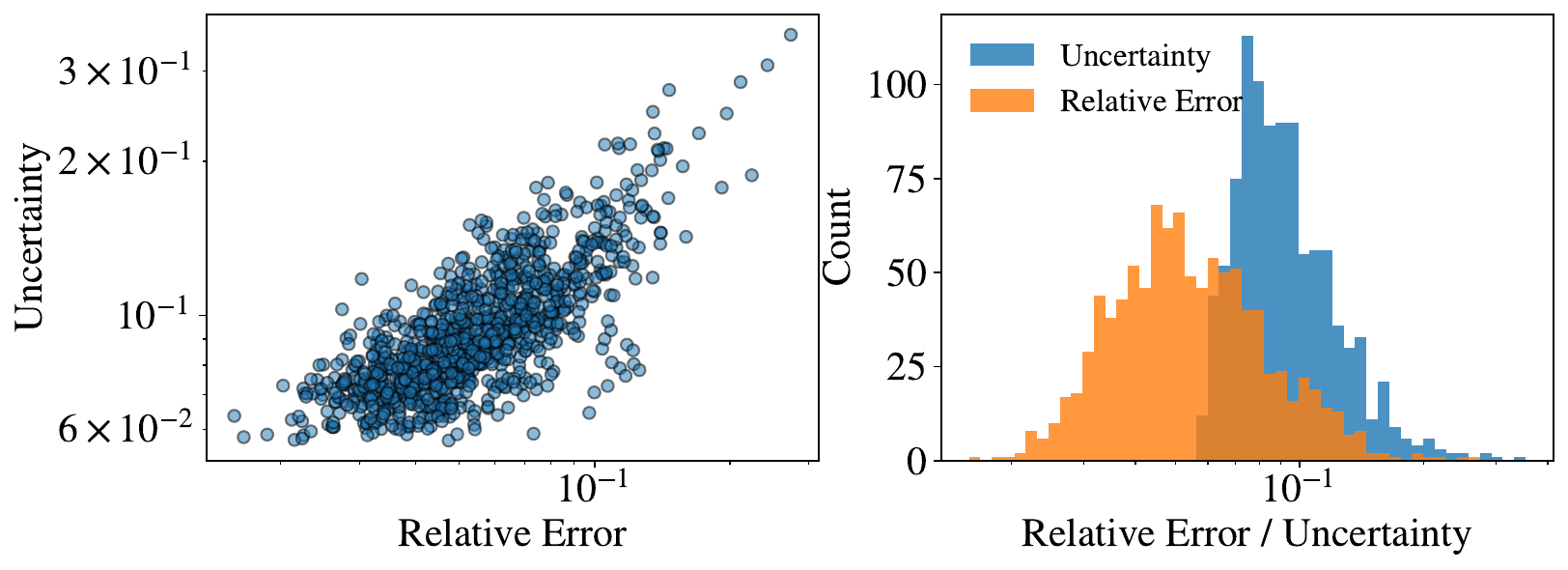}
\caption{Example \ref{subsec:ex2} large noise: \emph{(Left)}  Uncertainty versus relative error for outputs from the EKI DeepONet. \emph{(Right)} Histograms of relative error and uncertainties of the testing sample  outputs.}
\label{fig:Ex_2_uq_large}
\end{figure}

Finally,  Figure \ref{fig:Ex_2_uq_large} shows a similar positive correlation between relative error and uncertainty as we observed in the small noise regime. Additionally,  the marginal distribution analysis indicates that the uncertainty distribution conservatively approximates the relative error distribution.

\subsection{Gravity Pendulum}
\label{subsec:ex3}

Next, we consider an  ODE describing the gravity pendulum  \cite{Lu_2021}:
\begin{align}
    \frac{d^2s}{dt^2} &= -\sin(s) + u, \quad t\in [0,1].
\end{align}
where $u$ is the input function and $s$ is the corresponding output function. The input functions $u$ are similarly generated from the same Gaussian process as in \eqref{al:gp_1}-\eqref{al:gp_2}. The input sensors $u$ utilized are placed along $m=100$ equally spaced input locations along the temporal domain. The output data $s$ is generated by solving the corresponding ODE at $100$ equally spaced query locations along the temporal domain.

\subsubsection{Small Noise}
For the gravity pendulum, we initially examine a scenario where the training data $s^l$ is corrupted by zero-mean Gaussian noise. The standard deviation of this noise is set to $1\%$ of the maximum absolute value of the output function $s^l$.

\begin{figure}[ht]
\centering
\subfloat[]{\includegraphics[width=.5\textwidth]{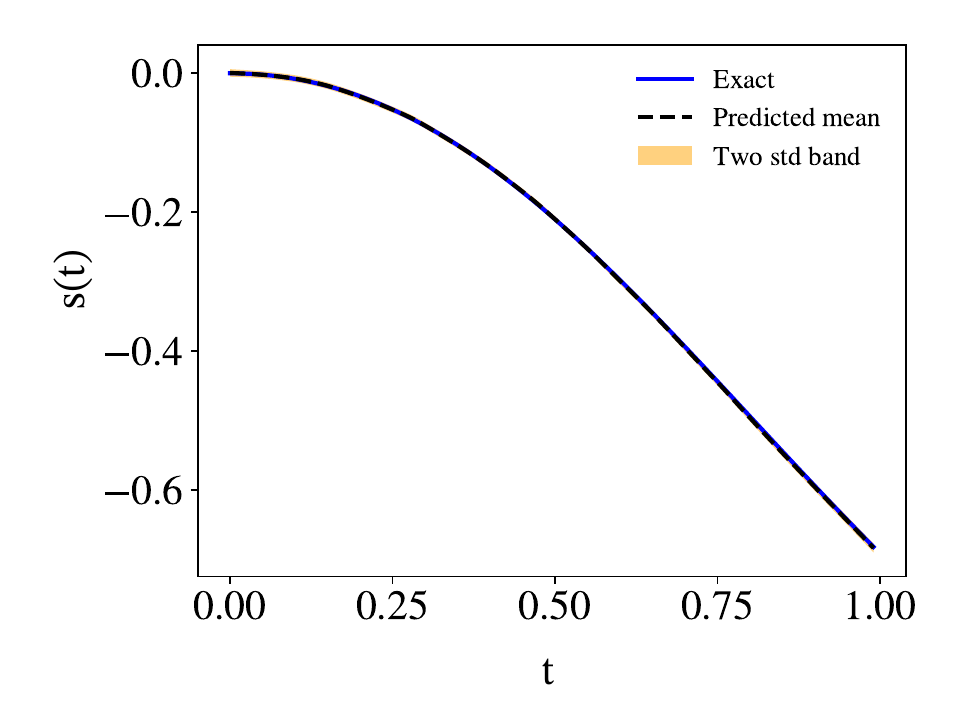}}
\hfill
\subfloat[]{\includegraphics[width=.5\textwidth]{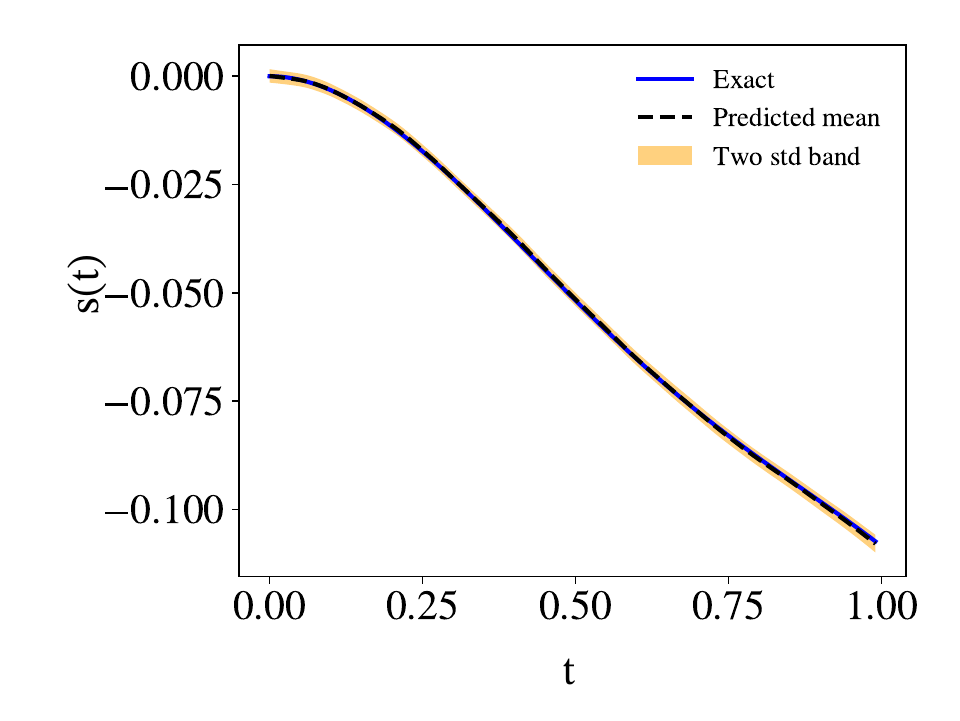}}\\
\subfloat[]{\includegraphics[width=.5\textwidth]{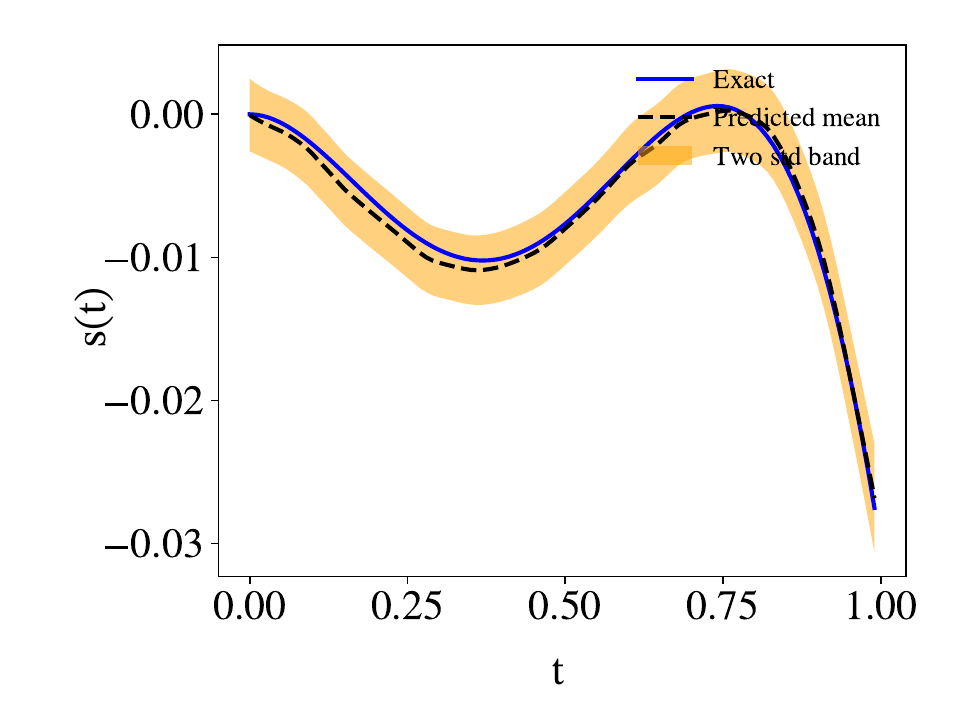}}
\caption{Example \ref{subsec:ex3} small noise: (a) and (b) two sample output function approximations with two standard deviation confidence intervals; (c) worst case scenario approximation of the output samples.}
\label{fig:ex_3_samples_small}
\end{figure}

In Figure \ref{fig:ex_3_samples_large}, we observe that the ensemble means in general well approximate the true solution in both representative and worst-case scenarios. Moreover, the two standard deviation confidence intervals encompass the true solutions in these instances. Examining Table \ref{tab:Ex_3_tab_small}, we find the average relative error of $0.8\%$ with the corresponding average uncertainty of $2.5\%$  across the entire testing dataset.  Furthermore, the nearly $99.9\%$ coverage suggests that the provided two standard-deviation confidence interval effectively captures the solution throughout the temporal domain for the majority of the test cases. Besides, the training time is roughly $214$ s for 572 EKI iterations.

\begin{table}[ht]
\centering
\begin{tabular}{|l|l|l|l|l|}
\hline
Relative Error & Uncertainty & Coverage & Training Time & Iterations\\ \hline
  0.008 & 0.025 &  0.999 & 214.4 Seconds & 572   \\ \hline
\end{tabular}
\caption{Example \ref{subsec:ex3} small noise: mean relative error, uncertainty, coverage metrics over the test dataset, training
  time, and the number of EKI iterations.  \label{tab:Ex_3_tab_small}}
\end{table}

\begin{figure}[ht]
\centering
\includegraphics[scale=0.45]{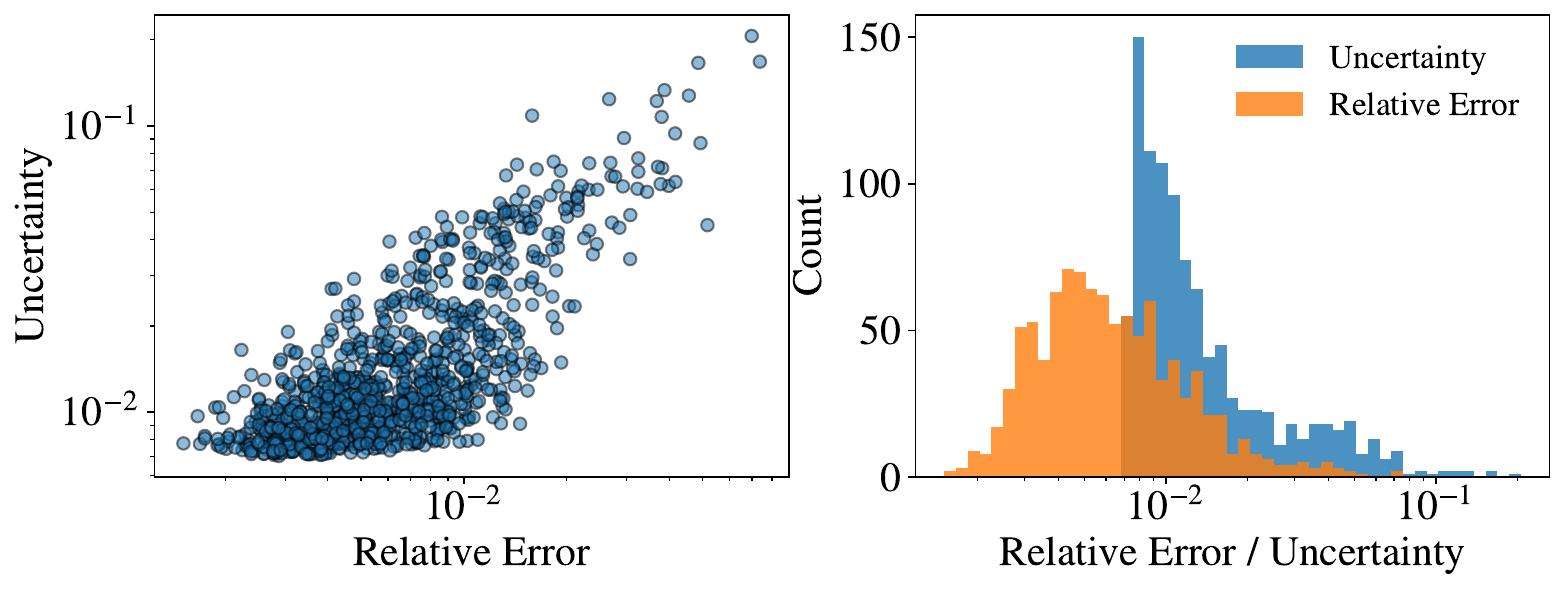}
\caption{Example \ref{subsec:ex3} small noise: \emph{(Left)}  Uncertainty versus relative error for outputs from the EKI DeepONet. \emph{(Right)} Histograms of relative error and uncertainties of the testing sample testing sample outputs.}
\label{fig:Ex_3_uq_small}
\end{figure}

Figure \ref{fig:Ex_3_uq_small} (left) clearly illustrates a strong positive correlation between relative error and uncertainty. Additionally, we see the largest outliers in the dataset in terms of relative error tend to exhibit a higher level of uncertainty. This correlation underscores the practical value of the uncertainty estimate as an informative indicator of error in the ensemble mean. Notably, for the examples with total relative error below $10^{-2}$, the uncertainty saturates around $6 \times 10^{-3}$. This is partially attributed to the uncertainty introduced during the perturbation of the prior ensemble members in each iteration. Consequently, there exists a lower bound beyond which the ensemble’s uncertainty cannot effectively characterize, potentially leading to an overestimation of uncertainty for the samples with more accurate prediction and, consequently, a conservative uncertainty quantification. To further confirm this,  the corresponding marginals are plotted in Figure \ref{fig:Ex_3_uq_small} (right), reinforcing that the uncertainty provides conservative estimates of the relative error distribution. 

\subsubsection{Large Noise}
We now consider the case where the output data set is corrupted by i.i.d. Gaussian noise, the standard deviation of which is set to $5\%$ of the maximum absolute value of each output function $ s^l$.

\begin{figure}[ht]
\centering
\subfloat[]{\includegraphics[width=.5\textwidth]{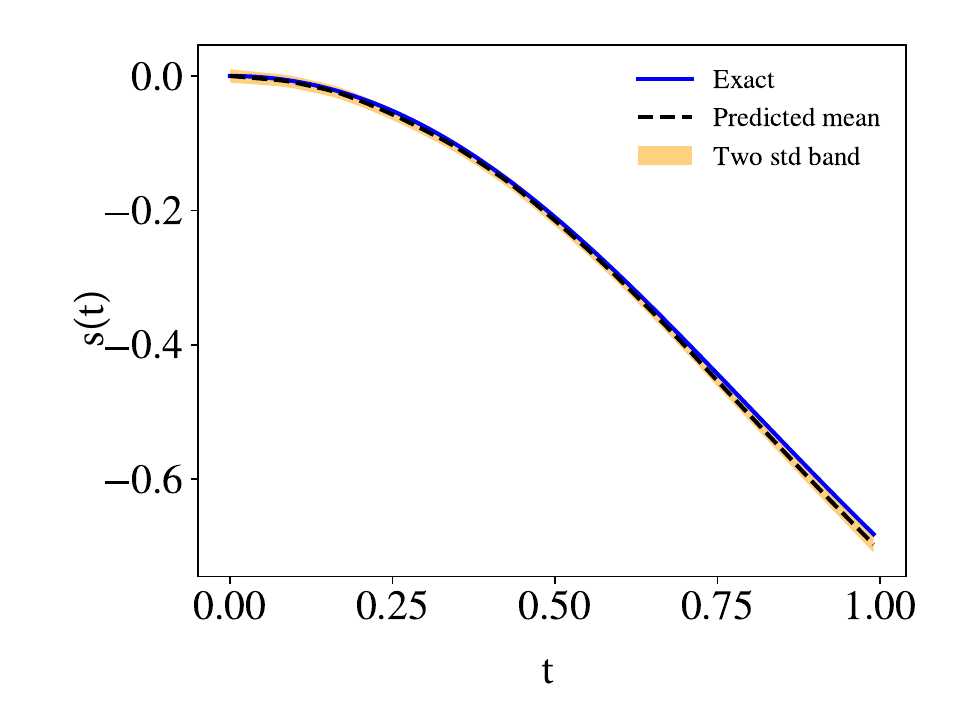}}
\hfill
\subfloat[]{\includegraphics[width=.5\textwidth]{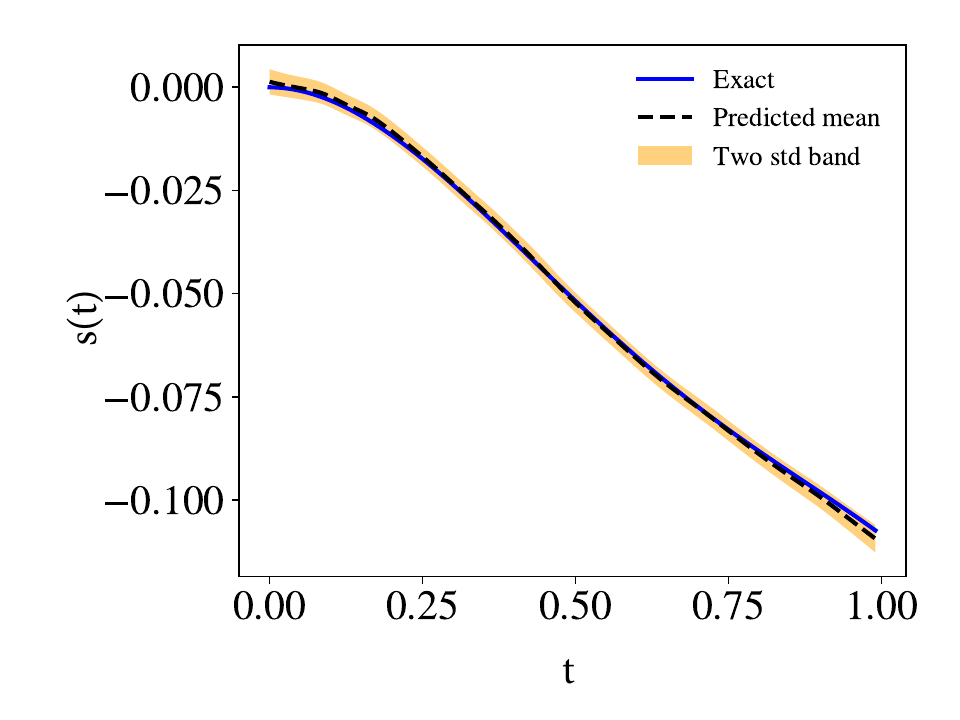}}\\
\subfloat[]{\includegraphics[width=.5\textwidth]{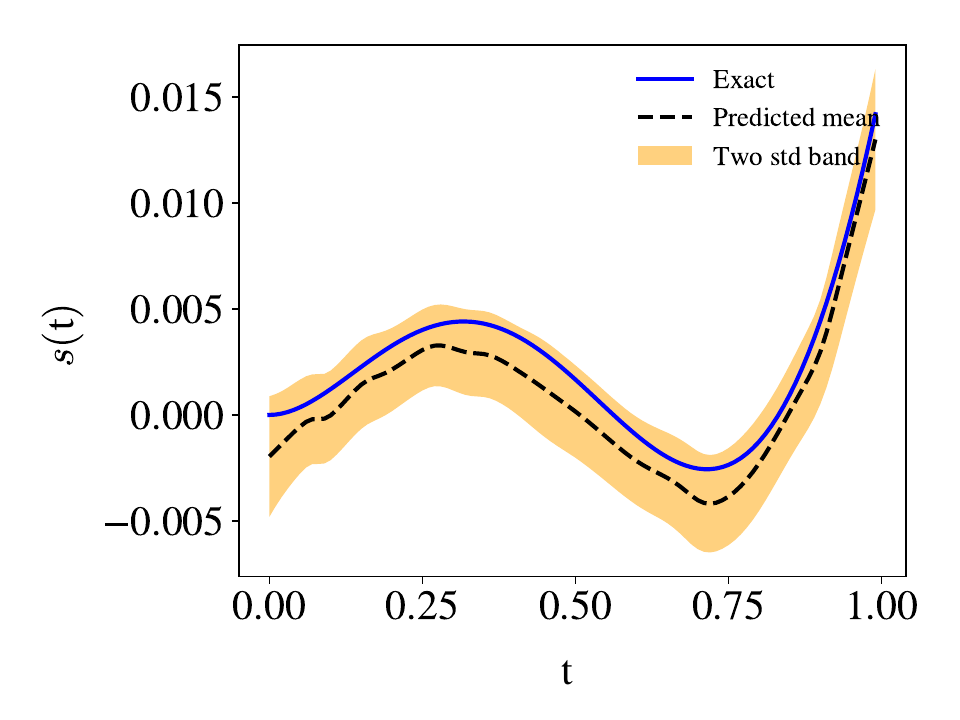}}
\caption{Example \ref{subsec:ex3} large noise: (a) and (b) two sample output function approximations with two standard deviation confidence intervals; (c) worst case scenario approximation of the output samples.}
\label{fig:ex_3_samples_large}
\end{figure}

For the large noise case, the posterior means seen in Figure \ref{fig:ex_3_samples_large} tend to be worse compared to the small noise scenario, for representative samples and worst-case scenarios.  Nonetheless, the corresponding confidence intervals are also larger and capable of capturing the true solution across the provided examples. Table \ref{tab:Ex_3_tab_large}, presents the average relative error over the test set is $1.9\%$, and the corresponding uncertainty is $3.6\%$. This observation reinforces the idea that in scenarios with higher noise levels, both error and uncertainty increase when compared to lower noise situations, which is the desired behavior. The resulting $96.8\%$ coverage of the solutions within the corresponding two standard deviation confidence interval indicates the uncertainty estimates provide reasonable estimates of the error in the corresponding solutions within the test datasets. Besides, the training time is roughly $256$ s for 686 EKI iterations.

\begin{table}[ht]
\centering
\begin{tabular}{|l|l|l|l|l|}
\hline
Relative Error & Uncertainty & Coverage & Training Time & Iterations\\ \hline
  0.019 & 0.036 &  0.968 & 255.8 Seconds & 686 \\ \hline
\end{tabular}
\caption{Example \ref{subsec:ex3} large noise: mean relative error, uncertainty, coverage metrics over the test dataset, training
  time, and the number of EKI iterations. \label{tab:Ex_3_tab_large} }
\end{table}

\begin{figure}[ht]
\centering
\includegraphics[scale=0.45]{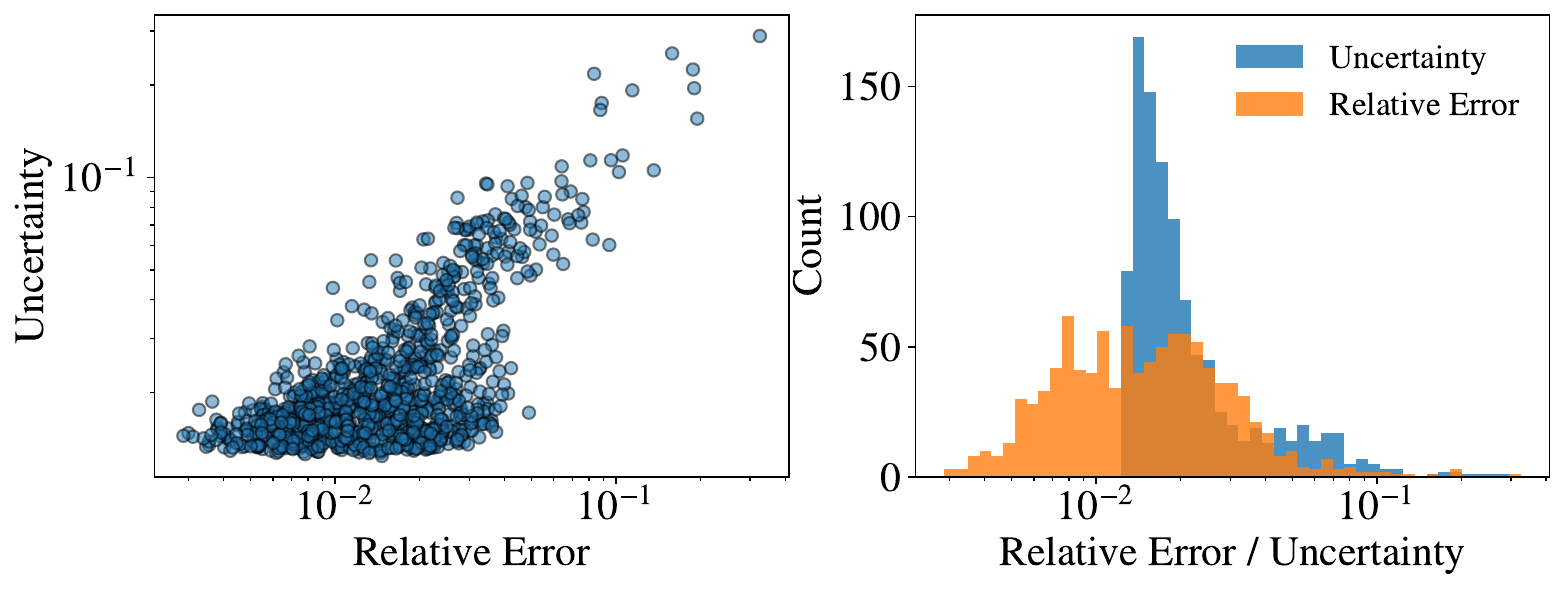}
\caption{Example \ref{subsec:ex3} large noise: \emph{(Left)}  Uncertainty versus relative error for outputs from the EKI DeepONet. \emph{(Right)} Estimates of marginal densities of relative error and uncertainties of the testing sample outputs.}
\label{fig:Ex_3_uq_large}
\end{figure}

In Figure \ref{fig:Ex_3_uq_large} (left), we observe a notable positive correlation between the relative error and the uncertainty. Specifically, more extreme outliers in the relative error tend to correspond to predictions with larger uncertainties. This relationship implies that the uncertainty can act as a valuable indicator of the reliability of the mean estimates produced by the DeepONets. Similar to the scenario with small noise, we observe an increasing non-linear relationship between relative error and uncertainty in log-log space, with a plateau of uncertainty for examples with a relative error below $5\times 10^{-3}$. 
In these cases, the uncertainty is bounded from below by $10^{-2}$, partially due to the uncertainty imposed by the prior ensemble perturbation in each EKI iteration. This potentially leads to an overestimation for most examples. Particularly, the uncertainty estimate for accurate samples by EKI-DeepONet tends to be more conservative. This is also consistent with the general trend found in the marginal distribution of the uncertainty in Figure \ref{fig:Ex_3_uq_large} (right), which tends to be more conservative when compared to the distribution of the relative error.

\section{Summary}
\label{sec:summary}
In this study, we proposed harnessing Ensemble Kalman Inversion (EKI) to efficiently train DeepONet ensembles and provide informative uncertainty estimates. To better handle large datasets, we adopt mini-batching techniques to mitigate computational demands. Additionally, we devised a heuristic approach to better estimate the covariance matrix of artificial dynamics to prevent the ensemble collapse while
minimizing excessive parameter perturbation during the training.  We evaluated the effectiveness of our approach across various benchmark problems at different noise levels.  Our EKI-based DeepONets demonstrated efficient inference capabilities with informative uncertainty estimates.

It's noteworthy that our study didn't explore scenarios involving very large networks. As network sizes expand, the demand for large ensemble sizes becomes essential to ensure the efficacy of EKI. However, accommodating these larger ensembles for extensive networks can lead to memory-intensive requirements. To address this challenge, we anticipate the potential application of dimension reduction techniques. Furthermore, localization techniques, which are well-established in the Ensemble Kalman Filter (ENKF) literature \cite{ott2004local, tong2023localized, ghattas2022non}, present an additional avenue for tackling the complexities associated with large ensembles for high dimensional problems. We plan to investigate these strategies in future work

\section*{Acknowledgments}
We thank for Investment in Strategic Priorities Program at the University of Iowa for their funding support.
\appendix
\section{Learning $Q$}
\label{sec:Q-learning}
In this section, we consider Example \ref{subsec:ex1}  to illustrate the benefits of learning the covariance matrix of the artificial dynamics described in Section \ref{subsec:Q}. We compare the results obtained from training the DeepONets using several fixed values of $Q$ with the adaptive learning method. Training output data  $\{s^l\}$ is corrupted by independent Gaussian noise with a standard deviation equal to 5\% of the maximum absolute value of the output function $s^l$ for each output function. 

\begin{figure}[ht]
\centering
\subfloat[]{\includegraphics[width=.5\textwidth]{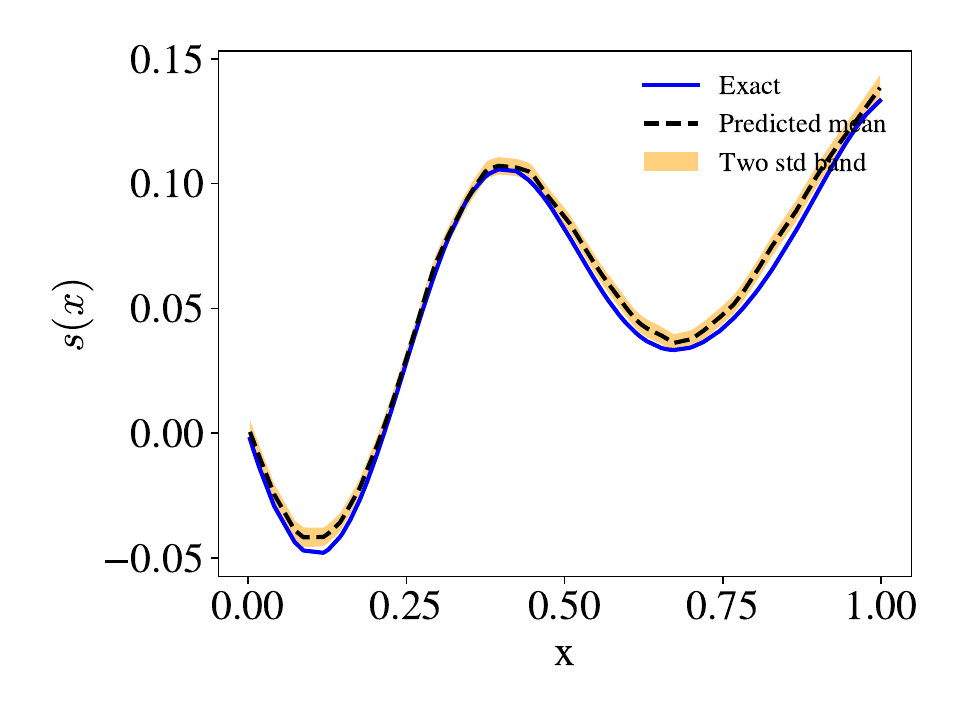}}
\hfill
\subfloat[]{\includegraphics[width=.5\textwidth]{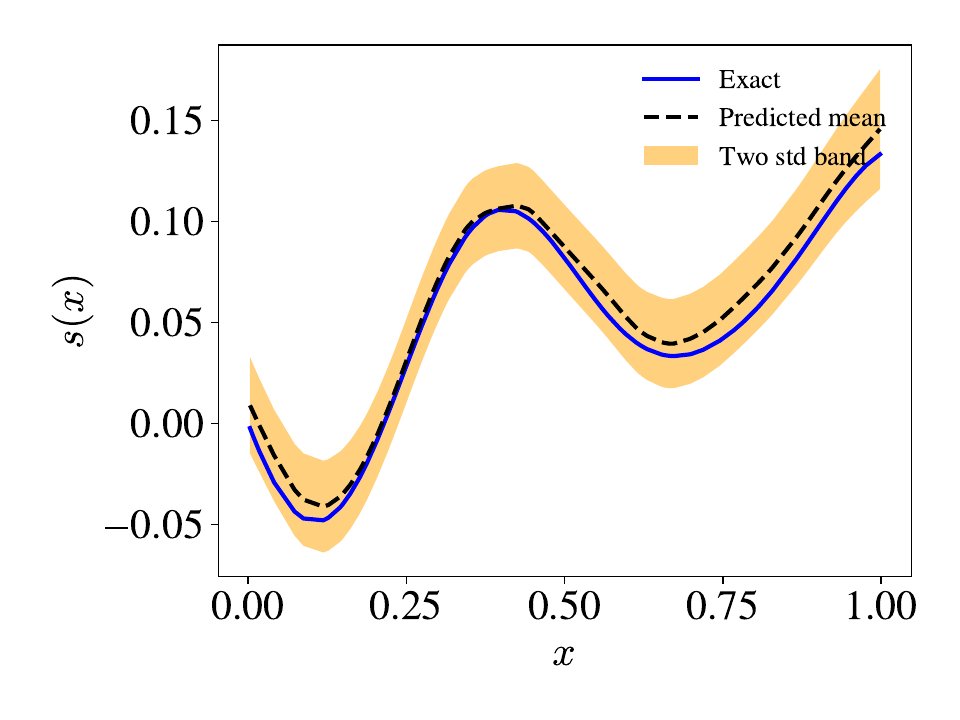}}\\
\subfloat[]{\includegraphics[width=.5\textwidth]{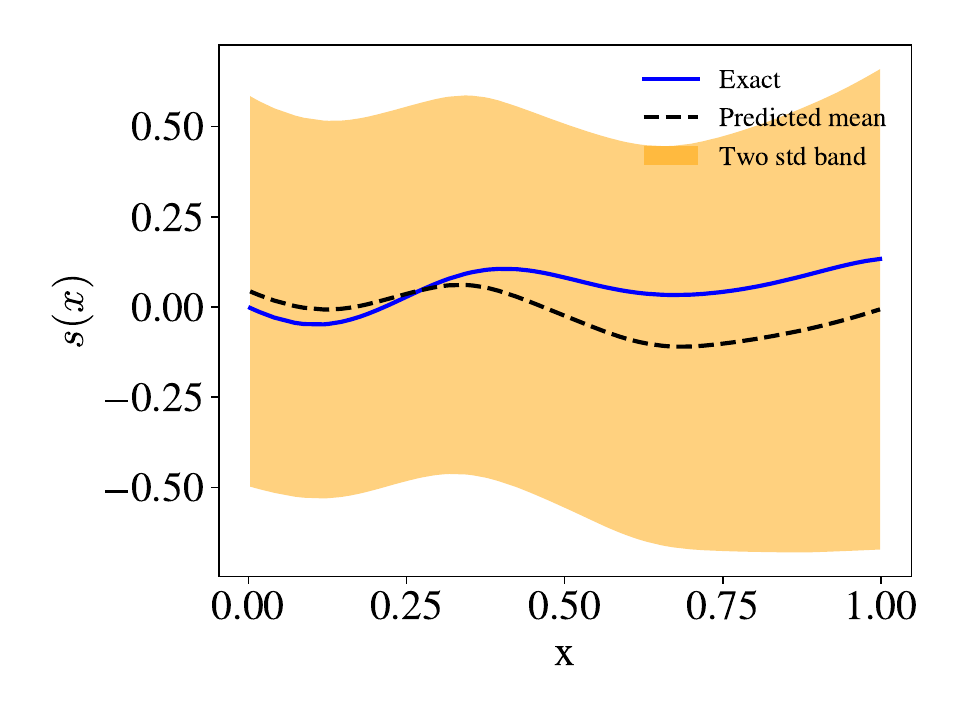}}
\hfill
\subfloat[]{\includegraphics[width=.5\textwidth]{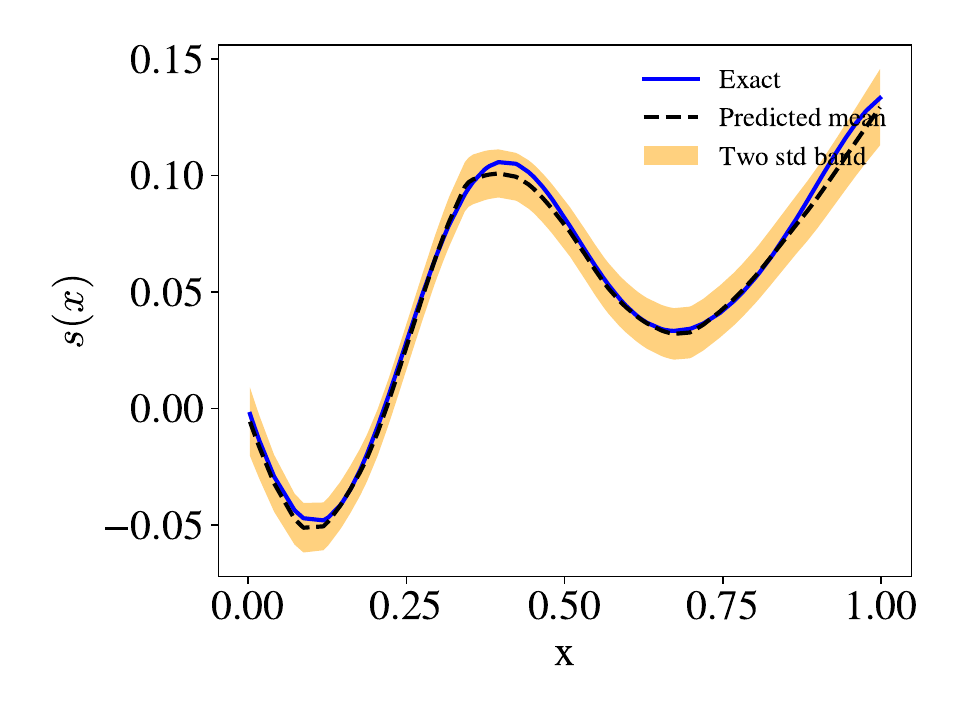}}\\
\caption{Example \ref{subsec:ex1} learning Q: a test sample with a) Fixed $Q=\sigma^2 I$ where $\sigma^2=0.0001$, b) Fixed $Q=\sigma^2 I$ where $\sigma^2=0.001$, c) Fixed $Q=\sigma^2 I$ where $\sigma^2=0.01$, d) Learned $Q$ where $Q_0=\sigma_0^2 I$ with $\sigma_0^2=0.0001$.}
\label{fig:ex_1_samples_Q}
\end{figure}

Figure \ref{fig:ex_1_samples_Q} shows a testing sample along with the ensemble mean and two standard deviation confidence interval  with different $Q=\sigma^2I$ .
The significance of $Q$ is evident in its impact on the mean and confidence interval. A larger $Q$ value leads to inaccurate mean estimates and overly conservative uncertainty bands, while a small $Q$ value results in overly confident confidence intervals. The adaptive $Q$ approach described in Section \ref{subsec:Q}, on the other hand, provides reasonable mean estimates and confidence intervals that reflect the mean prediction error.  For this example, the final value of $Q$ learned is $Q=\sigma^2I$ where $\sigma^2=0.0009$. 
In the rest of the section \ref{sec:NumExample}, we shall adopt the adaptive $Q$ approach for all examples.

\clearpage

\bibliographystyle{abbrv}
\bibliography{biblio}

\begin{thebibliography}{10}

\bibitem{Blei_2017}
D.~M. Blei, A.~Kucukelbir, and J.~D. McAuliffe.
\newblock Variational inference: A review for statisticians.
\newblock {\em Journal of the American Statistical Association},
  112(518):859--877, apr 2017.

\bibitem{cao2023lno}
Q.~Cao, S.~Goswami, and G.~E. Karniadakis.
\newblock Lno: Laplace neural operator for solving differential equations,
  2023.

\bibitem{chen2019neural}
R.~T.~Q. Chen, Y.~Rubanova, J.~Bettencourt, and D.~Duvenaud.
\newblock Neural ordinary differential equations, 2019.

\bibitem{chen1995universal}
T.~Chen and H.~Chen.
\newblock Universal approximation to nonlinear operators by neural networks
  with arbitrary activation functions and its application to dynamical systems.
\newblock {\em IEEE Transactions on Neural Networks}, 6(4):911--917, 1995.

\bibitem{Leoni_2021}
P.~C. Di~Leoni, L.~Lu, C.~Meneveau, G.~Karniadakis, and T.~A. Zaki.
\newblock Deeponet prediction of linear instability waves in high-speed
  boundary layers, 2021.

\bibitem{evensen2003ensemble}
G.~Evensen.
\newblock The ensemble kalman filter: Theoretical formulation and practical
  implementation.
\newblock {\em Ocean dynamics}, 53(4):343--367, 2003.

\bibitem{Shailesh_2023}
S.~Garg and S.~Chakraborty.
\newblock Vb-deeponet: A bayesian operator learning framework for uncertainty
  quantification.
\newblock {\em Engineering Applications of Artificial Intelligence},
  118:105685, 2023.

\bibitem{ghattas2022non}
O.~A. Ghattas and D.~Sanz-Alonso.
\newblock Non-asymptotic analysis of ensemble kalman updates: effective
  dimension and localization.
\newblock {\em arXiv preprint arXiv:2208.03246}, 2022.

\bibitem{guo2023ib}
L.~Guo, H.~Wu, W.~Zhou, and T.~Zhou.
\newblock Ib-uq: Information bottleneck based uncertainty quantification for
  neural function regression and neural operator learning.
\newblock {\em arXiv preprint arXiv:2302.03271}, 2023.

\bibitem{Guthweissman_2020}
P.~A. Guth, C.~Schillings, and S.~Weissmann.
\newblock Ensemble kalman filter for neural network based one-shot inversion.
\newblock {\em ArXiv}, abs/2005.02039, 2020.

\bibitem{houtekamer2001sequential}
P.~L. Houtekamer and H.~L. Mitchell.
\newblock A sequential ensemble kalman filter for atmospheric data
  assimilation.
\newblock {\em Monthly Weather Review}, 129(1):123--137, 2001.

\bibitem{huang2022iterated}
D.~Z. Huang, T.~Schneider, and A.~M. Stuart.
\newblock Iterated kalman methodology for inverse problems.
\newblock {\em Journal of Computational Physics}, 463:111262, 2022.

\bibitem{iglesias_2015}
M.~A. Iglesias.
\newblock Iterative regularization for ensemble data assimilation in reservoir
  models.
\newblock {\em Computational Geosciences}, 19(1):177--212, 2015.

\bibitem{Iglesias_2016}
M.~A. Iglesias.
\newblock A regularizing iterative ensemble kalman method for {PDE}-constrained
  inverse problems.
\newblock {\em Inverse Problems}, 32(2):025002, jan 2016.

\bibitem{Iglesias_2013}
M.~A. Iglesias, K.~J.~H. Law, and A.~M. Stuart.
\newblock Ensemble kalman methods for inverse problems.
\newblock {\em Inverse Problems}, 29(4):045001, mar 2013.

\bibitem{Kobyzev_2021}
I.~Kobyzev, S.~J. Prince, and M.~A. Brubaker.
\newblock Normalizing flows: An introduction and review of current methods.
\newblock {\em {IEEE} Transactions on Pattern Analysis and Machine
  Intelligence}, 43(11):3964--3979, nov 2021.

\bibitem{kovachki2022neural}
N.~Kovachki, Z.~Li, B.~Liu, K.~Azizzadenesheli, K.~Bhattacharya, A.~Stuart, and
  A.~Anandkumar.
\newblock Neural operator: Learning maps between function spaces, 2022.

\bibitem{Kovachki_2019}
N.~B. Kovachki and A.~M. Stuart.
\newblock Ensemble kalman inversion: a derivative-free technique for machine
  learning tasks.
\newblock {\em Inverse Problems}, 35(9):095005, aug 2019.

\bibitem{li2021fourier}
Z.~Li, N.~Kovachki, K.~Azizzadenesheli, B.~Liu, K.~Bhattacharya, A.~Stuart, and
  A.~Anandkumar.
\newblock Fourier neural operator for parametric partial differential
  equations, 2021.

\bibitem{Lin_2021}
C.~Lin, Z.~Li, L.~Lu, S.~Cai, M.~Maxey, and G.~E. Karniadakis.
\newblock Operator learning for predicting multiscale bubble growth dynamics.
\newblock {\em The Journal of Chemical Physics}, 154(10):104118, mar 2021.

\bibitem{lin2021accelerated}
G.~Lin, C.~Moya, and Z.~Zhang.
\newblock Accelerated replica exchange stochastic gradient langevin diffusion
  enhanced bayesian deeponet for solving noisy parametric pdes.
\newblock {\em arXiv preprint arXiv:2111.02484}, 2021.

\bibitem{liu2023dropout}
S.~Liu, S.~Reich, and X.~T. Tong.
\newblock Dropout ensemble kalman inversion for high dimensional inverse
  problems, 2023.

\bibitem{LopezGomez_2022}
I.~Lopez-Gomez, C.~Christopoulos, H.~L. Langeland~Ervik, O.~R.~A. Dunbar,
  Y.~Cohen, and T.~Schneider.
\newblock Training physics-based machine-learning parameterizations with
  gradient-free ensemble kalman methods.
\newblock {\em Journal of Advances in Modeling Earth Systems},
  14(8):e2022MS003105, 2022.
\newblock e2022MS003105 2022MS003105.

\bibitem{Lu_2021}
L.~Lu, P.~Jin, G.~Pang, Z.~Zhang, and G.~E. Karniadakis.
\newblock Learning nonlinear operators via {DeepONet} based on the universal
  approximation theorem of operators.
\newblock {\em Nature Machine Intelligence}, 3(3):218--229, mar 2021.

\bibitem{MAO2021110698}
Z.~Mao, L.~Lu, O.~Marxen, T.~A. Zaki, and G.~E. Karniadakis.
\newblock Deepm\&mnet for hypersonics: Predicting the coupled flow and
  finite-rate chemistry behind a normal shock using neural-network
  approximation of operators.
\newblock {\em Journal of Computational Physics}, 447:110698, 2021.

\bibitem{moya2022deeponet}
C.~Moya, S.~Zhang, M.~Yue, and G.~Lin.
\newblock Deeponet-grid-uq: A trustworthy deep operator framework for
  predicting the power grid's post-fault trajectories.
\newblock {\em arXiv preprint arXiv:2202.07176}, 2022.

\bibitem{ott2004local}
E.~Ott, B.~R. Hunt, I.~Szunyogh, A.~V. Zimin, E.~J. Kostelich, M.~Corazza,
  E.~Kalnay, D.~Patil, and J.~A. Yorke.
\newblock A local ensemble kalman filter for atmospheric data assimilation.
\newblock {\em Tellus A: Dynamic Meteorology and Oceanography}, 56(5):415--428,
  2004.

\bibitem{Pensoneault2023}
A.~Pensoneault and X.~Zhu.
\newblock Efficient bayesian physics informed neural networks for inverse
  problems via ensemble kalman inversion, 2023.

\bibitem{Prechelt2012}
L.~Prechelt.
\newblock {\em Early Stopping --- But When?}, pages 53--67.
\newblock Springer Berlin Heidelberg, Berlin, Heidelberg, 2012.

\bibitem{RAISSI2019686}
M.~Raissi, P.~Perdikaris, and G.~Karniadakis.
\newblock Physics-informed neural networks: A deep learning framework for
  solving forward and inverse problems involving nonlinear partial differential
  equations.
\newblock {\em Journal of Computational Physics}, 378:686--707, 2019.

\bibitem{tong2023localized}
X.~Tong and M.~Morzfeld.
\newblock Localized ensemble kalman inversion.
\newblock {\em Inverse Problems}, 39(6):064002, 2023.

\bibitem{TRIPURA2023115783}
T.~Tripura and S.~Chakraborty.
\newblock Wavelet neural operator for solving parametric partial differential
  equations in computational mechanics problems.
\newblock {\em Computer Methods in Applied Mechanics and Engineering},
  404:115783, 2023.

\bibitem{yang2022scalable}
Y.~Yang, G.~Kissas, and P.~Perdikaris.
\newblock Scalable uncertainty quantification for deep operator networks using
  randomized priors.
\newblock {\em Computer Methods in Applied Mechanics and Engineering},
  399:115399, 2022.

\bibitem{zou2023uncertainty}
Z.~Zou, X.~Meng, and G.~E. Karniadakis.
\newblock Uncertainty quantification for noisy inputs-outputs in
  physics-informed neural networks and neural operators.
\newblock {\em arXiv preprint arXiv:2311.11262}, 2023.

\end{thebibliography}

\end{document}